\documentclass[final,number]{elsarticle}

\usepackage[left=2cm, right=2cm, top=2cm]{geometry}
\usepackage[colorinlistoftodos]{todonotes}
\usepackage{rotating, amsmath,multirow,multicol,graphicx,subfigure}
\usepackage[noend]{algpseudocode}
\usepackage{algorithm}
\usepackage[colorlinks]{hyperref}
\hypersetup{pdftitle={Effects of Distance Measure Choice on KNN Classifier Performance - A Review}, colorlinks=true, citecolor=blue, linkcolor=red}

\newtheorem{Thm}{Theorem}[section]
\newtheorem{example}[Thm]{Example}
\makeatletter
\def\ps@pprintTitle{%
   \let\@oddhead\@empty
   \let\@evenhead\@empty
   \def\@oddfoot{\reset@font\hfil\thepage\hfil}
   \let\@evenfoot\@oddfoot
}
\makeatother

\begin{document}

\begin{frontmatter}

\title{Effects of Distance Measure Choice on KNN Classifier Performance - A Review}

\author[add1,add2,add3,add4]{V. B. Surya Prasath\corref{mycorrespondingauthor}}
\address[add1]{Division of Biomedical Informatics, Cincinnati Children's Hospital Medical Center, OH 45229 USA}
\address[add2]{Department of Pediatrics, University of Cincinnati College of Medicine, Cincinnati, OH USA}
\address[add3]{Department of Biomedical Informatics, College of Medicine, University of Cincinnati, OH 45267 USA}
\address[add4]{Department of Electrical Engineering and Computer Science, University of Cincinnati, OH 45221 USA}
\cortext[mycorrespondingauthor]{Corresponding author. Tel.: +1 513 636 2755}
\ead{prasatsa@uc.edu}

\author[mysecondaryaddress]{Haneen Arafat Abu Alfeilat}
\author[mysecondaryaddress]{Ahmad B. A. Hassanat}
\author[mysecondaryaddress]{Omar Lasassmeh}
\address[mysecondaryaddress]{Department of Information Technology, Mutah University, Karak, Jordan}
\author[third]{Ahmad S. Tarawneh}
\address[third]{Department of Algorithm and Their Applications, Eötvös Loránd University, Budapest, Hungary}

\author[address41,address42]{Mahmoud Bashir Alhasanat}
\address[address41]{Department of Geomatics, Faculty of Environmental Design. King Abdulaziz University, Jeddah, Saudi Arabia}
\address[address42]{Faculty of Engineering, Al-Hussein Bin Talal University, Maan, Jordan}

\author[mysecondaryaddress]{Hamzeh S. Eyal Salman}

\begin{abstract}
The K-nearest neighbor (KNN) classifier is one of the simplest and most common classifiers, yet its performance competes with the most complex classifiers in the literature. The core of this classifier depends mainly on measuring the distance or similarity between the tested examples and the training examples. This raises a major question about which distance measures to be used for the KNN classifier among a large number of distance and similarity measures available? This review attempts to answer this question through evaluating the performance (measured by accuracy, precision and recall) of the KNN using a large number of distance measures, tested on a number of real-world datasets, with and without adding different levels of noise. The experimental results show that the performance of KNN classifier depends significantly on the distance used, and the results showed large gaps between the performances of different distances. We found that a recently proposed non-convex distance performed the best when applied on most datasets comparing to the other tested distances. In addition, the performance of the KNN with this top performing distance degraded only about $20\%$ while the noise level reaches $90\%$, this is true for most of the distances used as well. This means that the KNN classifier using any of the top $10$ distances tolerate noise to a certain degree. Moreover, the results show that some distances are less affected by the added noise comparing to other distances.
\end{abstract}


\end{frontmatter}

\section{Introduction}\label{intro}

Classification is an important problem in big data, data science and machine learning. The K-nearest neighbor (KNN) is one of the oldest, simplest and accurate algorithms for patterns classification and regression models. KNN was proposed in 1951 by~\cite{Fix1951}, and then modified by~\cite{Cover1967}. KNN has been identified as one of the top ten methods in data mining~\citep{Wu2008}. Consequently, KNN has been studied over the past few decades and widely applied in many fields~\citep{Bhatia2010}. Thus, KNN comprises the baseline classifier in many pattern classification problems such as pattern recognition~\citep{Xu2008}, text categorization~\citep{Manne2012}, ranking models~\citep{Xiubo2008}, object recognition~\citep{Bajramovic2006}, and event recognition~\citep{Yang2000} applications.
KNN is a non-parametric algorithm~\cite{Kataria2013}. Non-Parametric means either there are no parameters or fixed number of parameters irrespective of size of data. Instead, parameters would be determined by the size of the training dataset. While there are no assumptions that need to be made to the underlying data distribution. Thus, KNN could be the best choice for any classification study that involves a little or no prior knowledge about the distribution of the data. In addition, KNN is one of the laziest learning methods. This implies storing all training data and waits until having the test data produced, without having to create a learning model~\cite{Wettschereck1997}. 

Despite its slow characteristic, surprisingly, KNN is used extensively for Big data classification, this includes the works of~\citep{Maillo2015, Maillo2017, Deng2016, Gallego2018, Wang2018}, this is due to the other good characteristics of the KNN such as simplicity and reasonable accuracy, since the speed issue is normally solved using some kind of divided-and-conquer approaches. Slowness is not the only problem associated with the KNN classifier, in addition to choosing the best K-neighbors problem~\citep{Hassanat2014b}, choosing the best distance/similarity measure is an important problem, this is because the performance of the KNN classifier is dependent on the distance/similarity measure used. This paper focuses on finding the best distance/similarity measure to be used with the KNN classifier to guarantee the highest possible accuracy.

\subsection{Related works}\label{ssec:rel}

Several studies have been conducted to analyze the performance of KNN classifier using different distance measures. Each study was applied  on various kinds of datasets with different distributions, types of data and using different numbers of  distance and similarity measures.

Chomboon et al~\citep{Chomboon2015} analyzed the performance of KNN classifier using 11 distance measures. These include Euclidean, Mahalanobis, Manhattan, Minkowski, Chebychev, Cosine, Correlation, Hamming, Jaccard, Standardized Euclidean and Spearman distances. Their experiment had been applied on eight binary synthetic datasets with various kinds of distributions that were  generated using MATLAB. They divided each dataset into 70\% for training set and 30\% for the testing set. The results showed that the Manhattan, Minkowski, Chebychev, Euclidean, Mahalanobis, and Standardized Euclidean distance measures achieved similar accuracy results and outperformed other tested distances.  

Punam and Nitin~\cite{Punam2015} evaluated the performance of KNN classifier using Chebychev, Euclidean, Manhattan, distance measures on KDD dataset~\citep{Tavallaee2009}. The KDD dataset contains 41 features and two classes which type of data is numeric. The dataset was normalized before conducting the experiment. To evaluate the performance of KNN, accuracy, sensitivity and  specificity  measures were calculated for each distance. The reported results indicate that the use of Manhattan distance outperform the other tested distances, with $97.8\%$ accuracy rate, $96.76\%$  sensitivity rate and $98.35\%$  Specificity rate. 

Hu et al~\cite{Hu2016} analyzed the effect of distance measures on KNN classifier for medical domain datasets. Their experiments were based on three different types of medical datasets containing categorical, numerical, and mixed types of data, which were chosen from the UCI machine learning repository, and four distance metrics including Euclidean, Cosine, Chi square, and Minkowsky distances. They divided each dataset into $90\%$ of data as training and $10\%$  as testing set, with K values from ranging from $1$ to $15$. The experimental results showed that Chi square distance function was the best choice for the three different types of datasets. However, using the Cosine, Euclidean and Minkowsky distance metrics performed the `worst' over the mixed type of datasets. The `worst' performance means the method with the lowest accuracy.

Todeschini et al~\cite{Todeschini2015,Todeschini2016} analyzed the effect of eighteen different distance measures on the performance of KNN classifier using eight benchmark datasets. The investigated distance measures included Manhattan, Euclidean, Soergel, Lance--Williams, contracted Jaccard--Tanimoto, Jaccard--Tanimoto, Bhattacharyya, Lagrange, Mahalanobis, Canberra, Wave-Edge, Clark, Cosine, Correlation and four Locally centered Mahalanobis distances. For evaluating the performance of these distances, the non-error rate and average rank were calculated for each distance. The result indicated that the `best' performance were the Manhattan, Euclidean, Soergel, Contracted Jaccard--Tanimoto and Lance--Williams distance measures. The `best' performance means the method with the highest accuracy.

Lopes and Ribeiro~\cite{Lopes2015} analyzed the impact of five distance metrics, namely Euclidean, Manhattan, Canberra, Chebychev and Minkowsky in instance-based learning algorithms. Particularly, 1-NN Classifier and the Incremental Hypersphere Classifier (IHC) Classifier, they reported the results of their empirical evaluation on fifteen datasets with different sizes showing that the Euclidean and Manhattan metrics significantly yield good results comparing to the other tested distances.

Alkasassbeh et al~\cite{Alkasassbeh2015} investigated the effect of Euclidean, Manhattan and a non-convex distance due to Hassanat~\cite{Hassanat2014a} distance  metrics on the performance of the KNN classifier, with K ranging from $1$ to the square root of the size of the training set, considering only the odd K's. In addition to experimenting on other classifiers such as the Ensemble Nearest Neighbor classifier (ENN)~\citep{Hassanat2014b}, and the Inverted Indexes of Neighbors Classifier (IINC)~\citep{Jirina2010}. Their experiments were conducted on $28$ datasets taken from the UCI machine learning repository, the reported results show that Hassanat distance~\cite{Hassanat2014a} outperformed both of Manhattan and Euclidean distances in most of the tested datasets using the three investigated classifiers. 

Lindi~\cite{Lindi2016} investigated three distance metrics to use the best performer among them with the KNN classifier, which was employed as a matcher for their face recognition system that was proposed for the NAO robot. The tested distances were Chi-square, Euclidean and Hassanat distances. Their experiments showed that Hassanat distance outperformed the other two distances in terms of precision, but was slower than both of the other distances.

\begin{table*}
\centering
	\caption{Comparison between previous studies for distance measures in KNN classifier along with `best' performing distance. Comparatively our current work compares the highest number of distance measures on variety of datasets.}\label{tab:previous}
	\begin{tabular}{llll}
	\hline
Reference		&	\#distances		&	\#datasets		&	Best distance\\
\hline
\cite{Chomboon2015} 	&	11	&	8	&	Manhattan, Minkowski\\
&&&Chebychev\\
&&&Euclidean, Mahalanobis\\
&&&Standardized Euclidean\\
\hline
\cite{Punam2015} 	&	3	&	1	&	Manhattan\\
\hline
\cite{Hu2016}	&	4	&	37	&	Chi square\\
\hline
\cite{Todeschini2015} 	&	18	&	8	&	Manhattan, Euclidean, Soergel\\
&&&Contracted Jaccard--Tanimoto\\
&&& Lance--Williams\\
\hline
\cite{Lopes2015} &	5	&	15	&	Euclidean and Manhattan\\
\hline
\cite{Alkasassbeh2015}	&	3	&	28	&	Hassanat\\
\hline
\cite{Lindi2016}		&	3	&	2	&	Hassanat\\
\hline
Ours		&	54	&	28	&	Hassanat\\
	\hline
	\end{tabular}
\end{table*}

Table~\ref{tab:previous} provides a summary of these previous works on evaluating various distances within KNN classifier, along with the best distance assesed by each of them. 
As can be seen from the above literature review of most related works, that all of the previous works have investigated either a small number of distance and similarity measures (ranging from $3$ to $18$ distances), a small number of datasets, or both.
\subsection{Contributions}\label{ssec:contri}

In KNN classifier, the distances between the test sample and the training data samples are identified by different measures. Therefore, distance measures play a vital role in determining the final classification output~\citep{Hu2016}. Euclidean distance is the most widely used distance metric in KNN classifications, however, only few studies examined the effect of different distance metrics on the performance of KNN, these used a small number of distances, a small number of datasets, or both.   
Such shortage in experiments does not prove which distance is the best to be used with the KNN classifier. 
Therefore, this review attempts to bridge this gap by testing a large number of distance metrics on a large number of different datasets, in addition to investigating the distance metrics that least affected by added noise.
 
The KNN classifier can deal with noisy data, therefore, we need to investigate the impact of choosing different distance measures on the KNN performance when classifying  a large number of real-world datasets, in addition to investigate which distance has the lowest noise implications.
There are two main research questions addressed in this review:
\begin{enumerate}
	\item What is the "best" distance metric to be implemented with the KNN classifier?
	\item What is the "best" distance metric to be implemented with the KNN classifier in the case of noise existence?
\end{enumerate}
We mean by the `best distance metric' (in this review) is the one that allows the KNN to classify test examples with the highest precision, recall and accuracy, i.e. the one that gives best performance of the KNN in terms of accuracy. 

The previous questions were partially answered by the aforementioned works, however, most of the reviewed research in this regard used a small number of distances/similarity measures and/or a small number of datasets. This work investigates the use of a relatively large number of distances and datasets, in order to draw more significant conclusions, in addition to reviewing a larger number of distances in one place.

\subsection{Organization}\label{ssec:organ}

We organized our review as follows. 
First in Section~\ref{sec:knn} we provide an introductory overview to KNN classification method and present its history, characteristics, advantages and disadvantages. We review the definitions of various distance measures used in conjunction with KNN. 
Section~\ref{sec:exp} explains the datasets that were used in classification experiments, the structure of the experiments model, and the performance evaluations measures. We present and discuss the results produced by the experimental framework.
Finally, Section~\ref{sec:conc} we provide the conclusions and possible future directions.     

\section{KNN and distance measures}\label{sec:knn}

\subsection{Brief overview of KNN classifier}\label{ssec:brief}

\begin{figure}
\centering
	\includegraphics[width= 5cm]{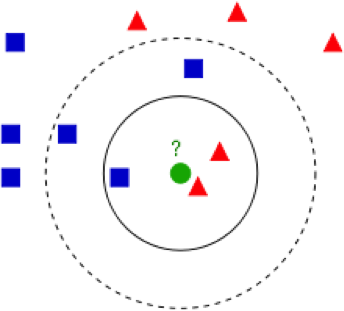}
	\caption{An example of  KNN classification with $K$ neighbors $K = 3$ (solid line circle) and $K = 5$ (dashed line circle), distance measure is Euclidean distance.}\label{fig1}
\end{figure}

The KNN algorithm classifies an unlabelled test sample based on the majority of similar samples among the k-nearest neighbors that are the closest to test sample. The distances between the test sample and each of the training data samples is determined by a specific distance measure. 
Figure~\ref{fig1} shows a KNN example, contains training samples with two classes, the first class is `blue square' and the second class is `red triangle'. The test sample is represented in green circle. These samples are placed into two dimensional feature spaces with one dimension for each feature. To classify the test sample that belongs to class `blue square' or to class `red triangle'; KNN adopts a distance function to find the K nearest neighbors to the test sample. Finding the majority of classes among the k nearest neighbors predicts the class of the test sample. In this case, when $k = 3$ the test sample is classified  to  the first class  `red triangle' because there are two  red triangles and only one  blue square inside the inner circle, but when $k = 5$ it is classified to the  "blue square" class because there are $2$ red triangles and only $3$ blue squares.

KNN is simple, but proved to be highly efficient and effective algorithm for solving various real life classification problems. However, KNN has got some disadvantages these include: 
\begin{enumerate}
	\item	How to find the optimum K value in KNN Algorithm?
	\item	High computational time cost as we need to compute the distance between each test sample to all training samples, for each test example we need $O(nm)$ time complexity (number of operations), where n is the number of examples in the training data, and m is the number of features for each example.
	\item	High Memory requirement as we need to store all training samples O(nm) space complexity.
	\item	Finally, we need to determine the distance function that is the core of this study. 
\end{enumerate}
The first problem was solved either by using all the examples and taking the inverted indexes~\citep{Jirina2008}, or using ensemble learning~\citep{Hassanat2014c}. For the second and third problems, many studies have proposed different solutions depending on reducing the size of the training dataset, those include and not limited to~\citep{Hart1968, Gates1972, Alpaydin1997, Kubat2000} and~\cite{Wilson2000}, or using approximate KNN classification such as~\citep{Arya1993} and~\citep{Zheng2016}. 
Although some previous studies in the literature that investigated the fourth problem (see Section~\ref{ssec:rel}), here we attempt to investigate the fourth problem on a much larger scale, i.e., investigating a large number of distance metrics tested on a large set of problems. In addition, we investigate the effect of noise on choosing the most suitable distance metric to be used by the KNN classifier.

\begin{algorithm}
\textbf{Input}: Training samples D, Test sample d, K \\
\textbf{Output}: Class label of test sample
\begin{algorithmic}[1]
\State   	Compute the distance between d and every sample in D
\State   	 Choose the K samples in D that are nearest to d; denote the set by P ($\in$D)
\State 	Assign d the class it that is the most frequent class (or the majority class) 
\end{algorithmic}
\caption{Basic KNN algorithm}\label{alg1} 
\end{algorithm}

The basic KNN classifier steps can be described as follows:
\begin{enumerate}
\item	Training phase: The training samples and the class labels of these samples are stored. no missing data allowed, no non-numeric data allowed.
\item	Classification phase: Each test sample is classified using majority vote of its neighbors by the following steps:
\begin{enumerate}
\item[a)]	Distances from the test sample to all stored training sample are calculated using a specific distance function or similarity measure. 
\item[b)]	The K nearest neighbors of the test sample are selected, where  K is a pre-defined small integer.
\item[c)]	The most repeated class of these K neighbors is assigned to the test sample. In other words, a test sample  is assigned to the class c if it is the most frequent class label among the K nearest training samples. If K = 1, then the class of the nearest neighbor is assigned to the test sample. KNN algorithm is described by Algorithm~\ref{alg1}.
\end{enumerate}
\end{enumerate}

We provide a toy example to illustrate how to compute the KNN classifier. Assuming that we have three training examples, having three attributes for each, and one test example as shown in Table~\ref{tab:train_examples}.

\begin{table}[ht]
\centering
	\caption{Training and testing data examples.}\label{tab:train_examples}
	\begin{tabular}{|l|l|l|l|l|}
\hline
&	X1	&	X2	&	X3	&	class\\
\hline
Training sample (1)	&	5	&	4	&	3	&	1\\
Training sample (2)	&	1	&	2	&	2	&	2\\
Training sample (3)	&	1	&	2	&	3	&	2\\
Test sample		&	4	&	4	&	2	&	?\\
\hline
	\end{tabular}
\end{table}

\textbf{Step1}: Determine the parameter K = number of  the nearest  neighbors to be considered. for this example we assume K = 1.

\textbf{Step 2}: Calculate the distance between test sample and all training samples using a specific similarity measure, in this example, Euclidean distance is used, see Table~\ref{tab:train_distances}.

\begin{table}[ht]
\centering
	\caption{Training and testing data examples with distances.}\label{tab:train_distances}
	\begin{tabular}{|l|l|l|l|l|l|}
\hline
&	X1	&	X2	&	X3	&	class		&	Distance\\
\hline
Training sample (1)	&	5	&	4	&	3	&	1	&	$D=\sqrt{(4-5)^2+(4-4)^2+(2-3)^2}=1.4$\\
\hline
Training sample (2)	&	1	&	2	&	2	&	2	&	$D=\sqrt{(4-1)^2+(4-2)^2+(2-2)^2}=3.6$\\
\hline
Training sample (3)	&	1	&	2	&	3	&	2	&	$D=\sqrt{(4-1)^2+(4-2)^2+(2-3)^2}=3.7$\\
\hline
Test sample		&	4	&	4	&	2	&	?	&\\
\hline
	\end{tabular}
\end{table}

\textbf{Step 3}: Sort all examples based on their similarity or distance to the tested example, and then keep only the K similar (nearest) examples as shown in Table~\ref{tab:train_distances_sorted}:

\begin{table}[ht]
\centering
	\caption{Training and testing data examples with distances.}\label{tab:train_distances_sorted}
	\begin{tabular}{|l|l|l|l|l|l|l|}
\hline
&	X1	&	X2	&	X3	&	class		&	Distance	&	Rank minimum distance\\
\hline
Training sample (1)	&	5	&	4	&	3	&	1	&	1.4	&	1\\
Training sample (2)	&	1	&	2	&	2	&	2	&	3.6	&	2\\
Training sample (3)	&	1	&	2	&	3	&	2	&	3.7	&	3\\
Test sample		&	4	&	4	&	2	&	?	&		&	\\
\hline
	\end{tabular}
\end{table}

\textbf{Step 4}: Based on the minimum distance, the class of the test sample is assigned to be $1$. However, if $K=3$ for instance, the class will be $2$.

\subsection{Noisy data}

The existence of noise in data is mainly related to the way that has been applied to acquire and preprocess data from its environment~\citep{Nettleton2010}. Noisy data is a corrupted form of data in some way, which leads to partial alteration of the data values.
Two main sources of noise can be identified: First, the implicit errors caused by measurement tools, such as using different types of sensors. Second, the random errors caused by batch processes or experts while  collecting data, for example, errors during the process document digitization. Based on these two sources of errors, two types of noise can be classified in a given dataset~\citep{Zhu2004}:
\begin{enumerate}
	\item Class noise: occurs when the sample is incorrectly labeled due to several causes such as data entry errors during labeling process, or the inadequacy of information that is being used to label each sample.
	\item	Attribute noise: refers to the corruptions in values of one or more attributes due to several causes, such as failures in sensor devices, irregularities in sampling or transcription errors~\citep{Garcia2014}. 
\end{enumerate}
The generation of noise can be classified by three main characteristics~\citep{Saez2013}:
\begin{enumerate}
	\item	The place where the noise is introduced: Noise may affect the attributes, class, training data, and test data separately or in combination. 
	\item	The noise distribution: The way in which the noise is introduced, for example, uniform or Gaussian. 
	\item	The magnitude of generated noise values: The extent to which the noise affects the data can be relative to each data value of each attribute, or relative to the standard deviation, minimum,  maximum for each attribute.
\end{enumerate}
In this work, we will add different noise levels to the tested datasets, to find the optimal distance metric that is least affected by this added noise with respect to the KNN classifier performance.

\subsection{Distance measures review}\label{ssec:dist}

The first appearance of the word distance can be found in the writings of Aristoteles (384 AC - 322 AC), who argued that   the word distance means: ``It is between extremities that distance is greatest" or ``things which have something between them, that is, a certain distance". In addition, ``distance has the sense of dimension [as in space has three dimensions, length, breadth and depth]". Euclid, one of the most important mathematicians of the ancient history, used the word distance only in his third postulate of the Principia~\citep{Euclid1956}: ``Every circle can be described by a centre and a distance". The distance is a numerical description of how far apart entities are. In data mining, the distance means a concrete way of describing what it means for elements of some space to be close to or far away from each other. Synonyms for distance include farness, dissimilarity, diversity, and synonyms for similarity include proximity~\citep{Cha2007}, nearness~\citep{Todeschini2015}. 

The distance function between two vectors x and y is a function $d(x,y)$ that defines the distance between both vectors as a non-negative real number.  This function is considered as a metric if satisfy a certain number of properties~\citep{Deza2009} that include the following:
\begin{enumerate}
	\item \textbf{Non-negativity}: The distance between x and y is always a value greater than or equal to zero.
	\[d(x,y) \geq 0 \]

	\item \textbf{Identity of indiscernibles}: The distance between x and y is equal to zero if and only if $x$ is equal to $y$.
	\[d(x,y)=0 \quad \text{iff} \quad x=y\]

	\item \textbf{Symmetry}: The distance between $x$ and $y$ is equal to the distance between y and x.
	\[d(x,y)=d(y,x)\]

	\item \textbf{Triangle inequality}: Considering the presence of a third point $z$, the distance between $x$ and $y$ is always less than or equal to the sum of the distance between $x$ and $z$ and the distance between $y$ and $z$. 
	\[d(x,y)\leq d(x,z)+d(z,y)\]
\end{enumerate}
When the distance is in the range $[0, 1]$, the calculation of a  corresponding similarity measure  $s(x,y)$ is as follows:
	\[s(x,y)=1-d(x,y)	\]

We consider the eight major distance families which consist of fifty four total distance measures. We categorized these distance measures following a similar categorization done by~\cite{Cha2007}. In what follows, we give the mathematical definitions of distances to measure the closeness between two vectors $x$ and $y$, where $x=(x_1, x_2,...,x_n)$ and $y = (y_1, y_2,...,y_n)$ having numeric attributes. As an example, we show the computed distance value between the example vectors $v1=\{5.1, 3.5, 1.4, 0.3\}$, $v2=\{5.4, 3.4, 1.7, 0.2\}$ as a result in each of these categories of distances reviewed here. Theoretical analysis of these various distance metrics is beyond the scope of this work.

\begin{enumerate}
	\item \textbf{$L_p$ Minkowski distance measures}:
	This family of distances includes three distance metrics that are special cases of Minkowski distance, corresponding to different values of  $p$ for this power distance. The Minkowski distance, which is also known as $L_p$ norm, is a generalized metric. It is defined as:
	\[D_{Mink}(x,y) = \sqrt[p]{\sum_{i=1}^n \vert{x_i - y_i}\vert^p},\]
	where $p$ is a positive value. When $p=2$, the distance becomes the Euclidean distance. When $p=1$ it becomes Manhattan distance. Chebyshev distance is a variant of Minkowski distance where $p = \infty$. $x_i$ is the $i^{\text{th}}$  value in the vector $x$ and  $y_i$  is the $i^{\text{th}}$  value in the vector $y$.
	
		\begin{enumerate}
			\item[1.1] Manhattan (MD): The Manhattan distance, also known as  $L_1$ norm, Taxicab norm, Rectilinear distance or City block distance, which considered by Hermann Minkowski in 19th-century Germany. This distance represents the sum of the absolute differences between the opposite values in vectors.
			\[MD(x,y) = \sum_{i=1}^n \vert{x_i - y_i}\vert\]
			
			\item[1.2] Chebyshev (CD): Chebyshev distance is also known as maximum value distance~\citep{Grabusts2011},  Lagrange~\citep{Todeschini2015} and chessboard distance~\citep{Premaratne2014}. This distance is appropriate in cases when two objects are to be defined as different if they are different in any one dimension~\citep{Verma2012}. It is a metric defined on a vector space where distance between two vectors is the greatest of their difference along any coordinate dimension. 
			\[CD(x,y) = \max_{i} \vert{x_i - y_i}\vert\]
			
			\item[1.3] Euclidean (ED): Also known as $L_2$ norm or Ruler distance, which is an extension to the Pythagorean Theorem. This distance represents the root of the sum of the square of  differences between the opposite values in vectors.			
			\[ED(x,y) = \sqrt{\sum_{i=1}^n \vert{x_i - y_i}\vert^2}\]
		\end{enumerate}

\begin{center}
	$L_p$ Minkowski distance measures\\
	\begin{tabular}{llll}
	\hline
	Abbrev.	&	Name	&	Definition	&	Result\\
	\hline
	MD	&	Manhattan	& $\sum_{i=1}^n \vert{x_i - y_i}\vert$		&0.8\\
	CD	&	Chebyshev 	&  $\max_{i} \vert{x_i - y_i}\vert$		&0.3\\
	ED	&	Euclidean 		& $\sqrt{\sum_{i=1}^n \vert{x_i - y_i}\vert^2}$	&0.4472\\
	\hline	
	\end{tabular}
\end{center}	
				
	\item \textbf{$L_1$ Distance measures}:
	This distance family depends mainly  finding the absolute difference, the family include Lorentzian, Canberra, Sorensen, Soergel, Kulczynski, Mean Character, Non Intersection distances.
	
			\begin{enumerate}
			\item[2.1] Lorentzian distance (LD): Lorentzian distance is represented by the natural log of the absolute difference between two vectors. This distance is sensitive to small changes since the log scale expands the lower range and compresses the higher range. 
			\[LD(x,y) = \sum_{i=1}^n ln(1+\vert x_i - y_i\vert),\]
			where $ln$ is the natural logarithm, and To ensure that the non-negativity property and to avoid log of zero, one is added.
			
			\item[2.2] Canberra distance (CanD): Canberra distance, which is introduced by~\cite{Williams1966} and modified in~\cite{Lance1967}. It is a weighted version of Manhattan distance, where the absolute difference between the attribute values of the vectors  $x$ and $y$ is divided by the sum of the absolute attribute values prior to summing~\citep{Akila2013}. This distance is mainly used for positive values.  It is very sensitive to small changes near zero, where it is more sensitive to proportional than to absolute differences. Therefore, this characteristic becomes more apparent in higher dimensional space, respectively with an increasing number of variables.  The Canberra distance is often used for data scattered around an origin.
			\[CanD(x,y) = \sum_{i=1}^n \frac{\vert x_i - y_i\vert}{\vert x_i \vert + \vert y_i \vert}\]
			
			\item[2.3] Sorensen distance (SD): The Sorensen distance~\citep{Sorensen1948}, also known as Bray--Curtis is one of the most commonly applied measurements to express relationships in ecology, environmental sciences and related fields. It is a modified Manhattan metric, where the summed differences between the attributes values of the vectors x and y are standardized by their summed attributes values~\citep{Szmidt2013}. When all the vectors values are positive, this measure take value between zero and one. 
			\[SD(x,y) = \frac{\sum_{i=1}^n\vert x_i - y_i\vert}{\sum_{i=1}^n(x_i +  y_i)}\]

			\item[2.4] Soergel distance (SoD): Soergel distance is one of the distance measures that is widely used to calculate the evolutionary distance~\citep{Chetty2008}. It is also known as Ruzicka distance.  For binary variables only, this distance is identical to the complement of the Tanimoto (or Jaccard) similarity coefficient~\citep{Zhou2008}. This distance obeys all four metric properties provided by all attributes have nonnegative values~\citep{Willett1998}.
			\[SoD(x,y) = \frac{\sum_{i=1}^n\vert x_i - y_i\vert}{\sum_{i=1}^n \max{(x_i,  y_i)}}\]
			
			\item[2.5] Kulczynski Distance (KD): Similar to the Soergel distance, but instead of using the maximum, it uses the minimum function.
			\[KD(x,y) = \frac{\sum_{i=1}^n\vert x_i - y_i\vert}{\sum_{i=1}^n \min{(x_i,  y_i)}}\]
			
			\item[2.6] Mean Character Distance (MCD): Also known as Average Manhattan, or Gower distance. 
			\[MCD(x,y) = \frac{\sum_{i=1}^n\vert x_i - y_i\vert}{n}\]
			
			\item[2.7]  Non Intersection Distance (NID): Non Intersection distance is the complement to the intersection similarity and is obtained by subtracting the intersection similarity from one. 
			\[NID(x,y) = \frac{1}{2}\sum_{i=1}^n \vert x_i - y_i \vert.\]
		\end{enumerate}

\begin{center}
	$L_1$ Distance measures\\	
	\begin{tabular}{llll}
	\hline
	Abbrev.	&	Name	&	Definition	&	Result\\
	\hline
	LD		&	 Lorentzian	&	$\sum_{i=1}^n ln(1+\vert x_i - y_i\vert)$	&	0.7153\\
	CanD	&	Canberra 	&	$\sum_{i=1}^n \frac{\vert x_i - y_i\vert}{\vert x_i \vert + \vert y_i \vert}$	&	0.0381\\
	SD		&	Sorensen	&	$ \frac{\sum_{i=1}^n\vert x_i - y_i\vert}{\sum_{i=1}^n(x_i +  y_i)}$	&	0.0381\\
	SoD		&	Soergel 	&	$ \frac{\sum_{i=1}^n\vert x_i - y_i\vert}{\sum_{i=1}^n \max{(x_i,  y_i)}}$	&0.0734\\
	KD		&	Kulczynski &	$ \frac{\sum_{i=1}^n\vert x_i - y_i\vert}{\sum_{i=1}^n \min{(x_i,  y_i)}}$	&	0.0792\\
	MCD		&	Mean Character &	$ \frac{\sum_{i=1}^n\vert x_i - y_i\vert}{n}$		&	0.2\\
	NID		&	Non Intersection &	$\frac{1}{2}\sum_{i=1}^n \vert x_i - y_i \vert$		&	0.4\\
	\hline
	\end{tabular}
\end{center}		
	
	\item \textbf{Inner product distance measures}:
	Distance measures belonging to this family are calculated by some products of pair wise values from both vectors, this type of distances includes: Jaccard, Cosine, Dice, Chord distances. 

		\begin{enumerate}
			\item[3.1] 	Jaccard distance (JacD): The Jaccard distance  measures dissimilarity between sample sets, it is a complementary to the Jaccard similarity coefficient~\citep{Jaccard1901} and is obtained by subtracting the Jaccard coefficient from one. This distance is a metric~\citep{Cesare2012}. 
			\[JacD(x,y) = \frac{\sum_{i=1}^n (x_i - y_i)^2}{\sum_{i=1}^n x_i^2 + \sum_{i=1}^n y_i^2 - \sum_{i=1}^n x_i y_i}\]
			
			\item[3.2] 	Cosine distance (CosD): The Cosine distance, also called angular distance, is derived from the cosine similarity that measures the angle between two vectors, where Cosine distance is obtained by subtracting the cosine similarity from one. 
			\[CosD(x,y) = 1- \frac{\sum_{i=1}^n x_i y_i}{\sqrt{\sum_{i=1}^n x_i^2} \sqrt{\sum_{i=1}^n y_i^2}}\]
			
			\item[3.3] 	Dice distance (DicD): The dice distance is derived from the dice similarity~\citep{Dice1945}, which is a complementary to the dice similarity and is obtained by subtracting the dice similarity from one. It can be sensitive to values near zero. This distance is a not a metric, in particular, the property of triangle inequality does not hold. This distance is widely used in information retrieval in documents and biological taxonomy.
			\[DicD(x,y) = 1- \frac{2\sum_{i=1}^n x_i y_i}{\sum_{i=1}^n x_i^2 + \sum_{i=1}^n y_i^2}\]
			
			\item[3.4] 	Chord distance (ChoD): A modification of Euclidean distance~\citep{Gan2007}, which was introduced by Orloci~\citep{Orloci1967} to be used in  analyzing  community composition data~\citep{Legendre2012}. It was defined as the length of the chord joining two normalized points within a hypersphere of radius one. This distance is one of the distance measures that is commonly used for clustering continuous data~\citep{Shirkhorshidi2015}.
			\[ChoD(x,y) = \sqrt{2- 2\frac{\sum_{i=1}^n x_i y_i}{\sum_{i=1}^n x_i^2  \sum_{i=1}^n y_i^2}}\]
		\end{enumerate}
\begin{center}
	Inner product distance measures family\\
	\begin{tabular}{llll}
	\hline
	Abbrev.	&	Name	&	Definition	&	Result\\
	\hline
	JacD		&	Jaccard		&	$\frac{\sum_{i=1}^n (x_i - y_i)^2}{\sum_{i=1}^n x_i^2 + \sum_{i=1}^n y_i^2 - \sum_{i=1}^n x_i y_i}$	& 0.0048\\
	CosD	&	Cosine		&	$1- \frac{\sum_{i=1}^n x_i y_i}{\sqrt{\sum_{i=1}^n x_i^2} \sqrt{\sum_{i=1}^n y_i^2}}$	&	0.0016\\
	DicD		&	Dice 			&	$1- \frac{2\sum_{i=1}^n x_i y_i}{\sum_{i=1}^n x_i^2 + \sum_{i=1}^n y_i^2}$	&	0.9524\\
	ChoD	&	Chord		&	$\sqrt{2- 2\frac{\sum_{i=1}^n x_i y_i}{\sum_{i=1}^n x_i^2  \sum_{i=1}^n y_i^2}}$	&	0.0564\\
	\hline
	\end{tabular}
\end{center}

	\item \textbf{Squared Chord distance measures}:
	Distances that belong to this family are obtained by calculating the sum of geometrics. The geometric mean of two values is the square root of their product. The distances in this family cannot be used with features vector of negative values, this family includes Bhattachayya, Squared Chord, Matusita, Hellinger distances.

		\begin{enumerate}
			\item[4.1] 	Bhattacharyya distance (BD): The Bhattacharyya distance measures the similarity of two probability distributions~\citep{Bhattachayya1943}.
			\[BD(x,y) = - ln \sum_{i=1}^n \sqrt{x_i y_i}\] 
			
			\item[4.2] 	Squared chord distance (SCD): Squared chord distance is mostly used with paleontologists and in studies on pollen. In this distance, the sum of square of square root difference at each point is taken along both vectors, which increases the difference for more dissimilar feature. 
			\[SCD(x,y) = \sum_{i=1}^n (\sqrt{x_i} -\sqrt{y_i})^2\]
			
			\item[4.3] Matusita distance (MatD): Matusita distance is the square root of the squared chord distance.
			\[MatD(x,y) = \sqrt{\sum_{i=1}^n (\sqrt{x_i} -\sqrt{y_i})^2}\]
			
			\item[4.4] 	Hellinger distance (HeD): Hellinger distance also called Jeffries - Matusita distance~\citep{Abbad2016} was introduced in 1909 by Hellinger~\citep{Hellinger1909}, it is a metric used to measure the similarity between two probability distributions. This distance is closely related to Bhattacharyya distance.
			\[HeD(x,y) = \sqrt{2\sum_{i=1}^n (\sqrt{x_i} -\sqrt{y_i})^2}\]
	
		\end{enumerate}	
\begin{center}	
	Squared Chord distance measures family\\
	\begin{tabular}{llll}
	\hline
	Abbrev.	&	Name	&	Definition	&	Result\\
	\hline	
	BD		&	Bhattacharyya 	&	$- ln \sum_{i=1}^n \sqrt{x_i y_i}$	&	-2.34996\\
	SCD		&	Squared Chord	&	$\sum_{i=1}^n (\sqrt{x_i} -\sqrt{y_i})^2$	&	0.0297\\
	MatD	&	Matusita 		&	$\sqrt{\sum_{i=1}^n (\sqrt{x_i} -\sqrt{y_i})^2}$	&	0.1722\\
	HeD		&	Hellinger 		&	$\sqrt{2\sum_{i=1}^n (\sqrt{x_i} -\sqrt{y_i})^2}$	&	0.2436 \\
	\hline
	\end{tabular}
\end{center}		
	
	\item \textbf{Squared $L_2$ distance measures}:
	In $L_2$ distance measure family, the square of difference at each point a long both vectors is considered for the total distance,  this family includes  Squared Euclidean, Clark, Neyman $\chi^2$, Pearson $\chi^2$, Squared $\chi^2$, Probabilistic Symmetric $\chi^2$, Divergence,  Additive Symmetric $\chi^2$, Average, Mean Censored Euclidean and Squared Chi-Squared distances.

		\begin{enumerate}
			\item[5.1] 	Squared Euclidean distance (SED): Squared Euclidean distance is the sum of the squared differences without taking the square root. 
			\[SED(x,y) = \sum_{i=1}^n (x_i - y_i)^2\]
			\setlength{\emergencystretch}{3em}
			\item[5.2] 	Clark distance (ClaD): The Clark distance also called coefficient of divergence was introduced by Clark~\citep{Clark1952}. It is the squared root of half of the divergence distance. 
			\[ClaD(x,y) = \sqrt{\sum_{i=1}^n \left(\frac{ x_i - y_i }{| x_i| +  |y_i| }\right)^2}\]
			
			\item[5.3] 	Neyman $\chi^2$ distance (NCSD): The Neyman $\chi^2$~\citep{Neyman1949} is called a quasi-distance.
			\[NCSD(x,y) = \sum_{i=1}^n \frac{(x_i - y_i)^2}{x_i}\] 
			
			\item[5.4] 	Pearson $\chi^2$ distance (PCSD): Pearson $\chi^2$ distance~\citep{Pearson1900}, also called $\chi^2$ distance. 
			\[PCSD(x,y) = \sum_{i=1}^n \frac{(x_i - y_i)^2}{y_i}\]
			
			\item[5.5] 	Squared $\chi^2$ distance (SquD): Also called triangular discrimination distance. This distance is a quasi-distance.
			\[SquD(x,y) = \sum_{i=1}^n \frac{(x_i - y_i)^2}{x_i+y_i}\]
			
			\item[5.6] 	Probabilistic Symmetric $\chi^2$ distance (PSCSD):  This distance is equivalent to Sangvi $\chi^2$ distance. 
			\[PSCSD(x,y) = 2\sum_{i=1}^n \frac{(x_i - y_i)^2}{x_i+y_i}\]
			
			\item[5.7] 	Divergence distance (DivD): 
			\[DivD(x,y) = 2\sum_{i=1}^n \frac{(x_i - y_i)^2}{(x_i+y_i)^2}\]
			
			\item[5.8] 	Additive Symmetric $\chi^2$ (ASCSD): Also known as symmetric $\chi^2$ divergence. 
			\[ASCSD(x,y) = 2\sum_{i=1}^n \frac{(x_i - y_i)^2(x_i+y_i)}{x_iy_i}\]
			
			\item[5.9] 	Average distance (AD): The average distance, also known as average Euclidean is a modified version of the Euclidean distance~\citep{Shirkhorshidi2015}. Where the Euclidean distance has the following  drawback, "if two data vectors have no attribute values in common, they may have a smaller distance than the other pair of data vectors containing the same attribute values"~\citep{Gan2007}, so that, this distance was adopted. 
			\[AD(x,y) = \sqrt{\frac{1}{n}\sum_{i=1}^n (x_i - y_i)^2}\]
			
			\item[5.10] Mean Censored Euclidean Distance (MCED): In this distance, the sum of squared differences between values is calculated and, to get the mean value, the summed value is divided by the total number of values where the pairs  values do not equal to zero. After that, the square root of the mean should be computed to get the final distance.
			\[MCED(x,y) = \sqrt{\frac{\sum_{i=1}^n (x_i - y_i)^2}{\sum_{i=1}^n 1_{x_i^2+y_i^2\neq 0}}}\]
			
			\item[5.11] Squared Chi-Squared (SCSD): 
			\[SCSD(x,y) = \sum_{i=1}^n \frac{(x_i - y_i)^2}{\vert x_i + y_i \vert}\]
		\end{enumerate}	
		
		
\begin{center}
	Squared $L_2$ distance measures family\\
	\begin{tabular}{llll}	
	\hline
	Abbrev.	&	Name	&	Definition	&	Result\\
	\hline
	SED		&	Squared Euclidean	&	$\sum_{i=1}^n (x_i - y_i)^2$	&	0.2\\
	ClaD		&	Clark 		&	$\sqrt{\sum_{i=1}^n \left(\frac{\vert x_i - y_i \vert}{x_i+y_i}\right)^2}$	&	0.2245\\
	NCSD	&	Neyman $\chi^2$	&	$\sum_{i=1}^n \frac{(x_i - y_i)^2}{x_i}$	&	0.1181\\
	PCSD	&	Pearson	$\chi^2$	& 	$\sum_{i=1}^n \frac{(x_i - y_i)^2}{y_i}$	&	0.1225\\
	SquD	&	Squared $\chi^2$	& 	$\sum_{i=1}^n \frac{(x_i - y_i)^2}{x_i+y_i}$	&	0.0591\\
	PSCSD	&	Probabilistic Symmetric $\chi^2$	& $2\sum_{i=1}^n \frac{(x_i - y_i)^2}{x_i+y_i}$	&	0.1182\\
	DivD		&	Divergence		&	$2\sum_{i=1}^n \frac{(x_i - y_i)^2}{(x_i+y_i)^2}$		&	0.1008\\
	ASCSD	&	Additive Symmetric $\chi^2$	& $2\sum_{i=1}^n \frac{(x_i - y_i)^2(x_i+y_i)}{x_iy_i}$	&	0.8054\\
	AD		&	Average 		&	$\sqrt{\frac{1}{n}\sum_{i=1}^n (x_i - y_i)^2}$	&	0.2236 \\
	MCED	&	Mean Censored Euclidean	&	$\sqrt{\frac{\sum_{i=1}^n (x_i - y_i)^2}{\sum_{i=1}^n 1_{x_i^2+y_i^2\neq 0}}}$	&	0.2236\\
	SCSD	&	Squared Chi-Squared	&	$\sum_{i=1}^n \frac{(x_i - y_i)^2}{\vert x_i + y_i \vert}$	&	0.0591\\
	\hline
	\end{tabular}
\end{center}

	\item \textbf{Shannon entropy distance measures}:
	The distance measures belonging to this family are related to the Shannon entropy~\citep{Shannon2001}. These distances include Kullback-Leibler, Jeffreys, K divergence, Topsoe, Jensen-Shannon, Jensen difference distances.

		\begin{enumerate}
			\item[6.1] 	Kullback-Leibler distance (KLD): Kullback-Leibler distance was introduced by~\cite{Kullback1951}, it is also known as KL divergence, relative entropy, or information deviation, which measures the difference between two probability distributions. This distance is not a metric measure, because it is not symmetric. Furthermore, it does not satisfy triangular inequality property, therefore it is called quasi-distance. Kullback-Leibler divergence has been used in several natural language applications such as for query expansion, language models, and categorization~\citep{Pinto2007}.
			\[KLD(x,y) = \sum_{i=1}^n x_i \, ln \frac{x_i}{y_i},\] 
			where $ln$ is the natural logarithm.
			
			\item[6.2] Jeffreys Distance (JefD): Jeffreys distance~\citep{Jeffreys1946}, also called J-divergence or KL2- distance, is a symmetric version of the Kullback-Leibler distance.
			\[JefD(x,y) = \sum_{i=1}^n (x_i -y_i) \, ln \frac{x_i}{y_i}\]
			
			\item[6.3] K divergence Distance (KDD): 
			\[KDD(x,y) = \sum_{i=1}^n x_i \, ln \frac{2x_i}{x_i + y_i}\]
			
			\item[6.4] Topsoe Distance (TopD): The Topsoe distance~\citep{Topsoe2000}, also called information statistics, is a symmetric version of the Kullback-Leibler distance. The Topsoe distance is twice the Jensen-Shannon divergence. This distance is not a metric, but its square root is a metric. 
			\[TopD(x,y) = \sum_{i=1}^n x_i \, ln \left(\frac{2x_i}{x_i + y_i}\right) + \sum_{i=1}^n y_i \, ln \left(\frac{2 y_i}{x_i + y_i}\right)\]
			
			\item[6.5] Jensen-Shannon Distance (JSD): Jensen-Shannon distance is the square root of the Jensen Shannon divergence. It is the half of the Topsoe distance which uses the average method to make the K divergence symmetric.
			\[JSD(x,y) = \frac{1}{2}\left[\sum_{i=1}^n x_i \, ln \left(\frac{2x_i}{x_i + y_i}\right) + \sum_{i=1}^n y_i \, ln \left(\frac{2 y_i}{x_i + y_i}\right)\right]\] 
			
			\item[6.6] Jensen difference distance (JDD): Jensen difference distance was introduced by~\cite{Sibson1969}.
			\[JDD(x,y) = \frac{1}{2}\left[\sum_{i=1}^n \frac{x_i \, ln x_i + y_i ln y_i}{2} - \left(\frac{x_i + y_i}{2}\right)\, ln \left(\frac{x_i + y_i}{2}\right)\right]\]
		\end{enumerate}

\begin{center}
	Shannon entropy distance measures family\\
	\begin{tabular}{llll}
	\hline
	Abbrev.	&	Name	&	Definition	&	Result\\
	\hline
	KLD		&	Kullback-Leibler 	&	$\sum_{i=1}^n x_i \, ln \frac{x_i}{y_i}$	&	-0.3402\\
	JefD		&	Jeffreys 			&	$\sum_{i=1}^n (x_i -y_i) \, ln \frac{x_i}{y_i}$	&	0.1184\\
	KDD		&	K divergence		&	$\sum_{i=1}^n x_i \, ln \frac{2x_i}{x_i + y_i}$	&	-0.1853\\
	TopD		&	Topsoe			&	$\sum_{i=1}^n x_i \, ln \left(\frac{2x_i}{x_i + y_i}\right) + \sum_{i=1}^n y_i \, ln \left(\frac{2 y_i}{x_i + y_i}\right)$	&	0.0323\\
	JSD		&	Jensen-Shannon	& 	$\frac{1}{2}\left[\sum_{i=1}^n x_i \, ln \frac{2x_i}{x_i + y_i} + \sum_{i=1}^n y_i \, ln \frac{2 y_i}{x_i + y_i}\right]$	&	0.014809\\
	JDD		&	Jensen difference	&	$\frac{1}{2}\left[\sum_{i=1}^n \frac{x_i \, ln x_i + y_i ln y_i}{2} - \left(\frac{x_i + y_i}{2}\right)\, ln \left(\frac{x_i + y_i}{2}\right)\right]$	&	0.0074\\
	\hline
	\end{tabular}
\end{center}

	\item \textbf{Vicissitude distance measures}:
	Vicissitude distance family consists of four distances, Vicis-Wave Hedges, Vicis Symmetric, Max Symmetric $\chi^2$ and Min Symmetric $\chi^2$ distances. These distances were generated from syntactic relationship for the aforementioned distance measures.
	
		\begin{enumerate}
			\item[7.1] 	Vicis-Wave Hedges distance (VWHD): The so-called "Wave-Hedges distance" has been applied to compressed image retrieval~\citep{Hatzigiorgaki2003}, content based video retrieval~\citep{Patel2012}, time series classification~\citep{Giusti2013},image fidelity~\citep{Macklem2002}, finger print recognition~\citep{Bharkad2011}, etc.. Interestingly, the source of the "Wave-Hedges" metric has not been correctly cited, and some of the previously mentioned resources allude to it incorrectly as~\cite{Hedges1976}. The source of this metric eludes the authors, despite best efforts otherwise. Even the name of the distance "Wave-Hedges" is questioned~\citep{Hassanat2014a}.
			\[VWHD(x,y) = \sum_{i=1}^n   \frac{\vert x_i - y_i\vert}{\min(x_i,y_i)}\]
			
			\item[7.2] Vicis symmetric distance (VSD): Vicis Symmetric distance is defined by three formulas, VSDF1, VSDF2, VSDF3 as the following
			\[VSDF1(x,y) = \sum_{i=1}^n   \frac{(x_i - y_i)^2}{\min(x_i,y_i)^2},\] 
			\[VSDF2(x,y) =\sum_{i=1}^n   \frac{(x_i - y_i)^2}{\min(x_i,y_i)},\] 
			\[VSDF3(x,y) = \sum_{i=1}^n   \frac{(x_i - y_i)^2}{\max(x_i,y_i)}\]		
			
			\item[7.3] 	Max symmetric $\chi^2$ distance (MSCD): 
			\[MSCD(x,y) = \max\left(\sum_{i=1}^n   \frac{(x_i - y_i)^2}{x_i}, \sum_{i=1}^n   \frac{(x_i - y_i)^2}{y_i}\right)\]
			
			\item[7.4] 	Min symmetric $\chi^2$ distance (MiSCSD): 
			\[MiSCSD(x,y) = \min\left(\sum_{i=1}^n   \frac{(x_i - y_i)^2}{x_i}, \sum_{i=1}^n   \frac{(x_i - y_i)^2}{y_i}\right)\]
		\end{enumerate}
	
\begin{center}	
	Vicissitude distance measures family\\
	\begin{tabular}{llll}
	\hline
	Abbrev.	&	Name	&	Definition	&	Result\\
	\hline
	VWHD	&	Vicis-Wave Hedges	&	$\sum_{i=1}^n   \frac{\vert x_i - y_i\vert}{\min(x_i,y_i)}$	&	0.8025\\
	&\\
	VSDF1		&	Vicis Symmetric1	&	$\sum_{i=1}^n   \frac{(x_i - y_i)^2}{\min(x_i,y_i)^2}$ &	0.3002\\
	VSDF2		&	Vicis Symmetric2	&	$\sum_{i=1}^n   \frac{(x_i - y_i)^2}{\min(x_i,y_i)}$	&	0.1349\\
	VSDF3		&	Vicis Symmetric3	&	$\sum_{i=1}^n   \frac{(x_i - y_i)^2}{\max(x_i,y_i)}$	&	0.1058\\
	MSCD	&	Max  Symmetric $\chi^2$	&	$\max\left(\sum_{i=1}^n   \frac{(x_i - y_i)^2}{x_i}, \sum_{i=1}^n   \frac{(x_i - y_i)^2}{y_i}\right)$	&	0.1225\\
	MiSCSD	&	Min  Symmetric $\chi^2$	&	$\min\left(\sum_{i=1}^n   \frac{(x_i - y_i)^2}{x_i}, \sum_{i=1}^n   \frac{(x_i - y_i)^2}{y_i}\right)$	&	0.1181\\
	\hline
	\end{tabular}
\end{center}		
	
	\item \textbf{Other distance measures}:
	These metrics exhibits distance measures utilizing multiple ideas or measures from previous distance measures, these include and not limited to   Average ($L_1$,$L_\infty$), Kumar-Johnson, Taneja, Pearson, Correlation, Squared Pearson, Hamming, Hausdorff, $\chi^2$ statistic, Whittaker's index of association, Meehl, Motyka and Hassanat distances.
	
		\begin{enumerate}
			\item[8.1] 	Average  ($L_1$, $L_\infty$) distance (AvgD): Average ($L_1$, $L_\infty$) distance is the average of Manhattan and Chebyshev distances.
			\[AvgD(x,y) = \frac{\sum_{i=1}^n \vert x_i - y_i\vert + \max_i \vert x_i - y_i\vert}{2}\]
			
			\item[8.2] 	Kumar- Johnson Distance (KJD): 
			\[KJD(x,y) = \sum_{i=1}^n  \left(\frac{(x_i^2 + y_i^2)^2}{2(x_iy_i)^{3/2}}\right)\]
			
			\item[8.3] 	Taneja Distance (TanD):~\citep{Taneja1995}
			\[TJD(x,y) = \sum_{i=1}^n  \left(\frac{x_i + y_i}{2}\right)\, ln \left(\frac{x_i + y_i}{2\sqrt{x_iy_i}}\right)\]
			
			\item[8.4] 	Pearson  Distance (PeaD): The Pearson distance is derived from the PearsonÕs correlation coefficient, which measures the linear relationship between two vectors~\citep{Fulekar2009}. This distance is obtained by subtracting the PearsonÕs correlation coefficient from one.
			\[PeaD(x,y) = 1-\frac{ \sum_{i=1}^n (x_i -  \bar x) (y_i - \bar y)}{\sqrt{\sum_{i=1}^n (x_i -  \bar x)^2 \star \sum_{i=1}^n (y_i - \bar y)^2}}\]
			where $\bar x = \frac{1}{n} \sum_{i=1}^n x_i$.
			
			\item[8.5] 	Correlation Distance (CorD): Correlation distance is a version of the Pearson distance, where the Pearson distance is scaled in order to obtain a distance measure in the range between zero and one.
			\[CorD(x,y) = \frac{1}{2}\left(1-\frac{ \sum_{i=1}^n (x_i -  \bar x) (y_i - \bar y)}{\sqrt{\sum_{i=1}^n (x_i -  \bar x)^2} \sqrt{\sum_{i=1}^n (y_i - \bar y)^2}}\right)\]
			
			\item[8.6] 	Squared Pearson Distance (SPeaD):
			\[SPeaD(x,y) = 1-\left(\frac{ \sum_{i=1}^n (x_i -  \bar x) (y_i - \bar y)}{\sqrt{\sum_{i=1}^n (x_i -  \bar x)^2} \sqrt{\sum_{i=1}^n (y_i - \bar y)^2}}\right)^2\]
			
			\item[8.7] 	Hamming Distance (HamD): Hamming distance~\citep{Hamming1958} is a distance metric that measures the number of mismatches between two vectors. It is mostly used for nominal data, string and bit-wise analyses, and also can be useful for numerical data.
			\[HamD(x,y) = \sum_{i=1}^n 1_{x_i\neq y_i}\]
			
			\item[8.8] 	Hausdorff  Distance (HauD):
			\[HauD(x,y) = \max(h(x,y),h(y,x))\] 
			where $h(x,y) = \max_{x_i\in x}\min_{y_i\in y} \vert\vert x_i - y_i\vert\vert$, and $\vert\vert\cdot\vert\vert$  is the vector norm (e.g. $L_2$  norm ). The function $h(x,y)$ is called the directed Hausdorff distance from $x$ to $y$. The Hausdorff distance HauD(x, y) measures the degree of mismatch between the sets $x$ and $y$ by measuring the remoteness between each point $x_i$ and $y_i$  and vice versa.
			
			\item[8.9] 	$\chi^2$ statistic Distance (CSSD): The $\chi^2$ statistic distance was used for image retrieval~\citep{Kadir2012}, histogram~\citep{Rubner2013}, etc. 
			\[CSSD(x,y) = \sum_{i=1}^n\frac{x_i - m_i}{m_i}\] 
			where $m_i = \frac{x_i+y_i}{2}$.
			
			\item[8.10] Whittaker's index of association Distance (WIAD): Whittaker's index of association distance was designed for species abundance data~\citep{Whittaker1952}.
			\[WIAD(x,y) = \frac{1}{2}\sum_{i=1}^n \vert \frac{x_i}{\sum_{i=1}^n x_i} - \frac{y_i}{\sum_{i=1}^n y_i}\vert\]
			
			\item[8.11] Meehl Distance (MeeD): Meehl distance depends on one consecutive point in each vector.
			\[MeeD(x,y) = \sum_{i=1}^{n-1} (x_i - y_i - x_{i+1} + y_{i+1})^2\]
			
			\item[8.12] Motyka Distance (MotD):
			\[MotD(x,y) = \frac{\sum_{i=1}^n \max(x_i,y_i)}{\sum_{i=1}^n (x_i+y_i)}\]

\begin{figure}
\centering
	\includegraphics[width=10cm, height=5cm]{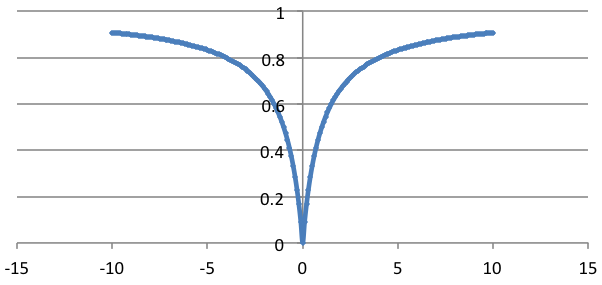}
	\caption{Representation of Hassanat distance between the point $0$ and $n$, where $n$ belongs to $[-10, 10]$.}\label{fig:hass}
\end{figure}	
			
			\item[8.13] Hassanat Distance (HasD): This is a non-convex distance introduced by~\cite{Hassanat2014a},
			\[HasD(x,y) = \sum_{i=1}^n D(x_i,y_i)\]
			where \[D(x,y) = \begin{cases}
						1 - \frac{1+\min(x_i,y_i)}{1+\max(x_i,y_i)},  & \min(x_i,y_i) \geq 0\\
						1 - \frac{1+\min(x_i,y_i) + \vert \min(x_i,y_i)\vert}{1+\max(x_i,y_i) + \vert \min(x_i,y_i)\vert},  & \min(x_i,y_i) < 0\\
						\end{cases}\]
		As can be seen, Hassanat distance is bounded by $[0,1[$. It reaches $1$ when the maximum value approaches infinity assuming the minimum is finite, or when the minimum value approaches minus infinity assuming the maximum is finite. This is shown by Figure~\ref{fig:hass} and the following equations.
		\[\lim_{\max(A_i,B_i)\to\infty} D(A_i,B_i) = \lim_{\max(A_i,B_i)\to -\infty} D(A_i,B_i) =1,\]
		By satisfying all the metric properties this distance was proved to be a metric by~\cite{Hassanat2014a}. In this metric no matter what the difference between two values is, the distance will be in the range of $0$ to $1$. so the maximum distance approaches to the dimension of the tested vectors, therefore the increases in dimensions increases the distance linearly in the worst case. 
		\end{enumerate}			
					
\end{enumerate}	

\newpage
				
\begin{center}
	Other distance measures family\\
	\begin{tabular}{llll}
	\hline
	Abbrev.	&	Name	&	Definition	&	Result\\
	\hline	
	AvgD	&	Average  ($L_1$, $L_\infty$) &	$\frac{\sum_{i=1}^n \vert x_i - y_i\vert + \max_i \vert x_i - y_i\vert}{2}$	&	0.55\\
	KJD		&	Kumar-Johnson	&	$\sum_{i=1}^n  \left(\frac{(x_i^2 + y_i^2)^2}{2(x_iy_i)^{3/2}}\right)$	&	21.2138\\
	TanD		&	Taneja 			&	$\sum_{i=1}^n  \left(\frac{x_i + y_i}{2}\right)\, ln \left(\frac{x_i + y_i}{2\sqrt{x_iy_i}}\right)$	&	0.0149\\
	PeaD	&	Pearson  			&	$1-\frac{ \sum_{i=1}^n (x_i -  \bar x) (y_i - \bar y)}{\sqrt{\sum_{i=1}^n (x_i -  \bar x)^2 \star \sum_{i=1}^n (y_i - \bar y)^2}}$	&\\
	&	&	 $\bar x = \frac{1}{n} \sum_{i=1}^n x_i$	&	0.9684\\
	CorD		&	Correlation 		&	$\frac{1}{2}\left(1-\frac{ \sum_{i=1}^n (x_i -  \bar x) (y_i - \bar y)}{\sqrt{\sum_{i=1}^n (x_i -  \bar x)^2} \sqrt{\sum_{i=1}^n (y_i - \bar y)^2}}\right)$	&	0.4842\\
	SPeaD	&	Squared Pearson 	&	$1-\left(\frac{ \sum_{i=1}^n (x_i -  \bar x) (y_i - \bar y)}{\sqrt{\sum_{i=1}^n (x_i -  \bar x)^2} \sqrt{\sum_{i=1}^n (y_i - \bar y)^2}}\right)^2$	&	0.999\\
	HamD	&	Hamming 			&	$\sum_{i=1}^n 1_{x_i\neq y_i}$	&	4\\
	HauD	&	Hausdorff  		&	$\max(h(x,y),h(y,x))$ &\\
	& 	&	$h(x,y) = \max_{x_i\in x}\min_{y_i\in y} \vert\vert x_i - y_i\vert\vert$	&	0.3\\
	CSSD	&	$\chi^2$	statistic 	&	$\sum_{i=1}^n\frac{x_i - m_i}{m_i}$, $m_i = \frac{x_i+y_i}{2}$	&	0.0894\\
	WIAD	&	Whittaker's index of assoc. 	&	$\frac{1}{2}\sum_{i=1}^n \vert \frac{x_i}{\sum_{i=1}^n x_i} - \frac{y_i}{\sum_{i=1}^n y_i}\vert$	&	1.9377\\
	MeeD	&	Meehl			&	$\sum_{i=1}^{n-1} (x_i - y_i - x_{i+1} + y_{i+1})^2$	&	0.48\\
	MotD	&	Motyka 			&	$\frac{\sum_{i=1}^n \max(x_i,y_i)}{\sum_{i=1}^n (x_i+y_i)}$	&	0.5190\\
	HasD	&	Hassanat			&	$\sum_{i=1}^n D(x_i,y_i)$ &\\ 
	&	& $ = \begin{cases}
						1 - \frac{1+\min(x_i,y_i)}{1+\max(x_i,y_i)},  & \min(x_i,y_i) \geq 0\\
						1 - \frac{1+\min(x_i,y_i) + \vert \min(x_i,y_i)\vert}{1+\max(x_i,y_i) + \vert \min(x_i,y_i)\vert},  & \min(x_i,y_i) < 0\\
						\end{cases}$	&	0.2571\\
	\hline
	\end{tabular}
\end{center}

\section{Experimental framework}\label{sec:exp}

\subsection{Datasets used for experiments}\label{ssec:data}

\begin{table}
\centering
	\caption{Description of the real-world datasets used (from the UCI Machine Learning Repository), where \#E means number of examples, and \#F means number of features and \#C means number of classes.}\label{tab:data}
	\begin{tabular}{lllllll}
	\hline
Name		&	\#E		&	\#F	&	\#C	&	data type		&	Min		&	Max\\
	\hline
Heart		&	270		&	25	&	2	&	real\& integer	&	0		&	564\\
Balance		&	625		&	4	&	3	&	positive integer	&	1		&	5\\
Cancer		&	683		&	9	&	2	&	positive integer	&	0		&	9\\
German		&	1000		&	24	&	2	&	positive integer	&	0		&	184\\
Liver			&	345		&	6	&	2	&	real\& integer	&	0		&	297\\
Vehicle		&	846		&	18	&	4	&	positive integer	&	0		&	1018\\
Vote			&	399		&	10	&	2	&	positive integer	&	0		&	2\\
BCW			&	699		&	10	&	2	&	positive integer	&	1		&	13454352\\
Haberman		&	306		&	3	&	2	&	positive integer	&	0		&	83\\
Letter rec.		&	20000	&	16	&	26	&	positive integer	&	0		&	15\\
Wholesale		&	440		&	7	&	2	&	positive integer	&	1		&	112151\\
Australian		&	690		&	42	&	2	&	positive real	&	0		&	100001\\
Glass		&	214		&	9	&	6	&	positive real	&	0		&	75.41\\
Sonar		&	208		&	60	&	2	&	positive real	&	0		&	1\\
Wine			&	178		&	13	&	3	&	positive real	&	0.13		&	1680\\
EEG			&	14980	&	14	&	2	&	positive real	&	86.67	&	715897\\
Parkinson		&	1040		&	27	&	2	&	positive real	&	0		&	1490\\
Iris			&	150		&	4	&	3	&	positive real	&	0.1		&	7.9\\
Diabetes		&	768		&	8	&	2	&	real\& integer	&	0		&	846\\
Monkey1		&	556		&	17	&	2	&	binary		&	0		&	1\\
Ionosphere	&	351		&	34	&	2	&	real			&	-1		&	1\\
Phoneme		&	5404		&	5	&	2	&	real			&	-1.82		&	4.38\\
Segmen		&	2310		&	19	&	7	&	real			&	-50		&	1386.33\\
Vowel		&	528		&	10	&	11	&	real			&	-5.21		&	5.07\\
Wave21		&	5000		&	21	&	3	&	real			&	-4.2		&	9.06\\
Wave40		&	5000		&	40	&	3	&	real			&	-3.97		&	8.82\\
Banknote		&	1372		&	4	&	2	&	real			&	-13.77	&	17.93\\
QSAR		&	1055		&	41	&	2	&	real			&	-5.256	&	147\\
	\hline
	\end{tabular}
\end{table}

The experiments were done on twenty eight datasets which represent real life classification problems, obtained from the UCI Machine Learning Repository~\citep{Lichman2013}. The UCI Machine Learning Repository is a collection of databases, domain theories, and data generators that are used by the machine learning community for the empirical analysis of machine learning algorithms. The database was created in $1987$ by David Aha and fellow graduate students at UC Irvine. Since that time, it has been widely used by students, educators, and researchers all over the world as a primary source of machine learning data sets. 

Each dataset consists of a set of examples. Each example is defined by a number of attributes and all the examples inside the data are represented by the same number of attributes. One of these attributes is called the class attribute, which contains the class value (label) of the data, whose values are predicted for test the examples. Short description of all the datasets used is provided in Table~\ref{tab:data}.
	
\subsection{Experimental setup}\label{ssec:frame}

\begin{figure}
\centering
	\includegraphics[width= 8cm, height=8cm]{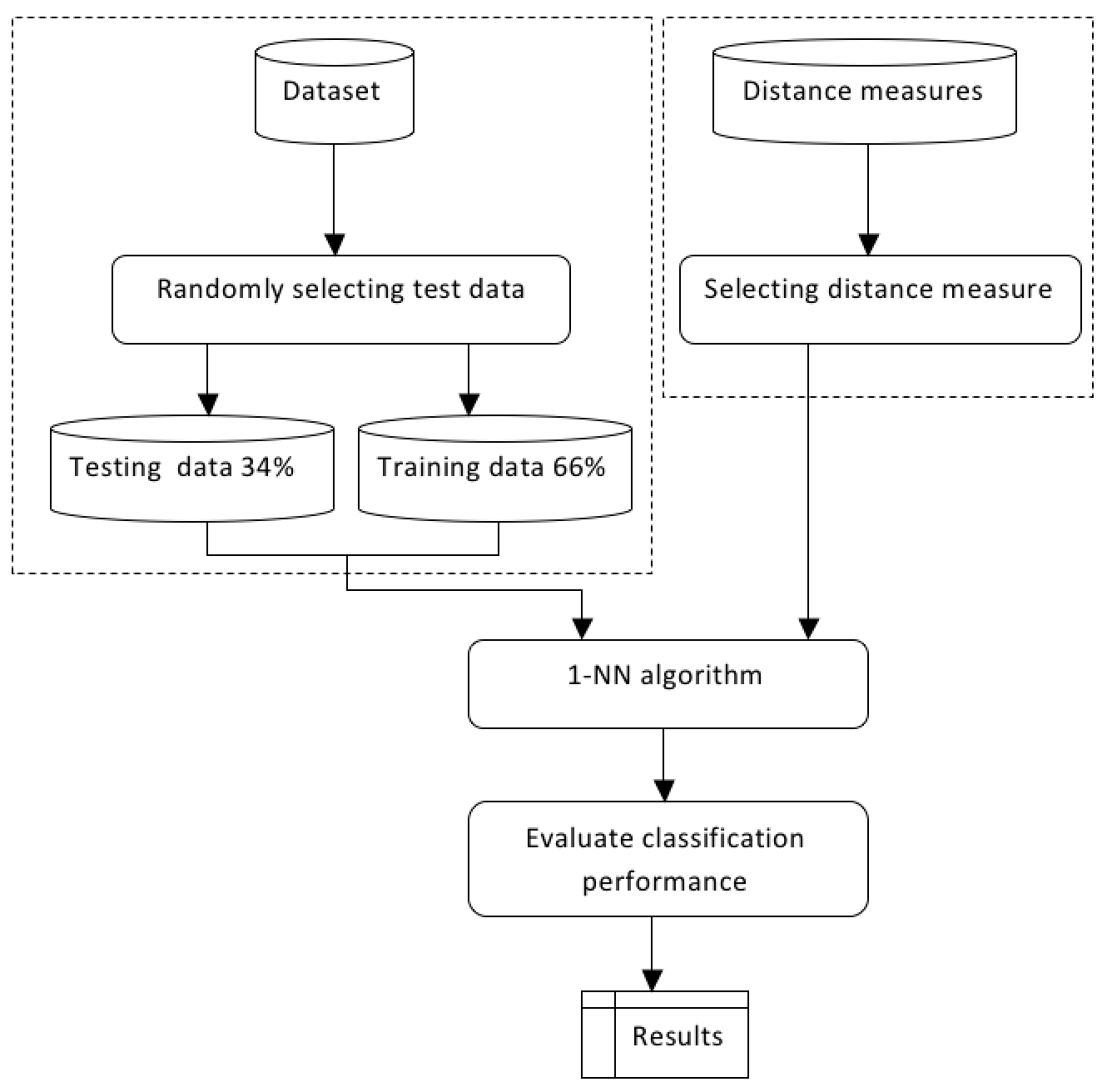}
	\caption{The framework of our experiments for discerning the effect of various distances on the performance of KNN classifier.}\label{fig:framework}
\end{figure}

\begin{algorithm}
\textbf{Input}: Original dataset $D$,  level of noise  $x\%$ [$10\%$-$90\%$]\\
\textbf{Output}: Noisy dataset
\begin{algorithmic}[1]
\State   	Number of noisy examples : $N = x\% *$ number of examples in $D$
\State   	 Array NoisyExample [$N$] 
\For{$K =1$ to $N$}
\State	            Randomly choose  an example number as  $E$ from $D$ 
\If	E is chosen previously
\State	Go to Step 4
\Else
\State	NoisyExample [$k$]$=E$
\EndIf
\EndFor
\For{each  attribute $A_i$} do
	\For{each NoisyExample  $NE_j$} do
\State	$RV$ = Random value between Min($A_i$) and Max($A_i$)
\State	$NE_j A_j = RV$.
	\EndFor
\EndFor
\end{algorithmic}
\caption{Create noisy dataset}\label{alg2} 
\end{algorithm}

Each dataset is divided into two data sets, one for training, and the other for testing. For this purpose, $34\%$ of the data set is used for testing, and $66\%$ of the data is dedicated for training. The value of K  is set to $1$ for simplicity. 
The $34\%$ of the data, which were used as a test sample, were chosen randomly, and each experiment on each data set was repeated $10$ times to obtain random examples for testing and training. The overall experimental framework is shown in Figure~\ref{fig:framework}. 
Our experiments are divided into two major parts:
\begin{enumerate}

\item The first part of experiments aims to find the best distance measures to be used by KNN classifier without any noise in the datasets. We used all the $54$ distances which were reviewed in Section~\ref{ssec:dist}. 

\item The second part of experiments aims to find the best distance measure to be used by KNN classifier in the case of noisy data. In this work, we define the `best' method as the method that performs with the highest accuracy. We added noise into each dataset at various levels of noise. The experiments in the second part were conducted using the top $10$ distances, those which achieved the best results in the first part of experiments. Therefore, in order to create a noisy dataset from the original one, a level of noise $x\%$ is selected in the range of ($10\%$ to $90\%$), the level of noise means the number of examples that need to be noisy, the amount of noise is selected randomly between the minimum and maximum values of each attribute, all attributes for each examples are corrupted by a random noise, the number of noisy examples are selected randomly. Algorithm~\ref{alg2} describes the process of corrupting data with random noise to be used for further experiments for the purposes of this work.
\end{enumerate}

\subsection{Performance evaluation measures}\label{ssec:perf}

Different measures are available for evaluating the performance of classifiers. In this study, three measures were used, accuracy, precision, and recall.
Accuracy is calculated to evaluate the overall classifier performance. It is defined as the ratio of   the test samples that are correctly classified to the number of tested examples,
\begin{eqnarray}\label{E:accuracy}	
Accuracy=  \frac{\text{Number of correct classifications}}{\text{Total number of test samples}}.
\end{eqnarray}
In order to assess the performance with respect to every class in a dataset, We compute precision and recall measures. Precision (or positive predictive value) is the fraction of retrieved instances that are relevant, while recall (or sensitivity) is the fraction of relevant instances that are retrieved. 
These measures can be constructed by computing the following:
\begin{enumerate}
\item True positive (TP): The number of correctly classified examples of a specific class (as we calculate these measures for each class)
\item	True negative (TN):The number of correctly classified examples that were not belonging to the specific class
\item	False positive (FP):The number of examples that incorrectly assigned to the specific class
\item	False negative (FN): The number of examples that incorrectly assigned to another class
\end{enumerate}

The precision and recall of a multi-class classification system are defined by,
\begin{eqnarray}\label{E:precision}
\text{Average Precision} = \frac{1}{N} \sum_{i=1}^N \frac{{TP}_i}{{TP}_i + {FP}_i},
\end{eqnarray}
\begin{eqnarray}\label{E:recall}
\text{Average Recall} = \frac{1}{N} \sum_{i=1}^N \frac{{TP}_i}{{TP}_i + {FN}_i},
\end{eqnarray}
where $N$ is the number of classes,  ${TP}_i$ is the number of true positive for class $i$, ${FN}_i$  is the number of false negative for class $i$ and ${FP}_i$  is the number of false positive for class $i$.

These performance measures can be derived from the confusion matrix. The confusion matrix is represented by a matrix that shows the predicted and actual classification. The matrix is $n\times n$, where $n$ is the number of classes. The structure of confusion matrix for multi-class classification is given by,
\begin{eqnarray}\label{E:confusion}
\left(
\begin{array}{c|cccc}
&& \text{Predicted Class}  &\\
&	\text{Classified as}~c_1	&	\text{Classified as}~c_{12} 	&	\cdots	&	\text{Classified as}~c_{1n}\\
\hline
\text{Actual Class}~c_1 & c_{11}		&	c_{12} 	&	\cdots	&	c_{1n}\\
\text{Actual Class}~c_2 & c_{21}		&	c_{22} 	&	\cdots	&	c_{2n}\\
\vdots 		  & \vdots		&	\vdots 	&	\vdots	&	\vdots\\
\text{Actual Class}~c_n & c_{n1}		&	c_{n2} 	&	\cdots	&	c_{nn}\\
\end{array}
\right)
\end{eqnarray}
This matrix reports the number of false positives, false negatives, true positives, and true negatives which are defined through elements of the confusion matrix as follows,
\begin{align}\label{E:fpfntp}
{TP}_i & =c_{ii} \\
{FP}_i & = \sum_{k=1}^N  c_{ki} - {TP}_i\\
{FN}_i &= \sum_{k=1}^N c_{ik} - {TP}_i\\
{TN}_i & = \sum_{k=1}^N\sum_{f=1}^N c_{kf} - {TP}_i - {FP}_i - {FN}_i
\end{align}
Accuracy, precision and recall will be calculated for the KNN classifier using all the  similarity measures and distance metrics discussed in Section~\ref{ssec:dist},  on all the datasets described in Table~\ref{tab:data}, this is to compare and asses the performance of the KNN classifier using different distance metrics and similarity measures.

\subsection{Experimental results and discussion}\label{sec:results}

For the purposes of this review, two sets of experiments have been conducted. The aim of the first set is to compare the performance of the KNN classifiers when used with each of the $54$ distances and similarity measures reviewed in Section~\ref{ssec:dist} without any noise. The second set of experiments is designed to find the most robust distance that affected the least with different noise levels.

\subsection{Without noise}\label{ssec:without}

A number of different predefined distance families were used in this set of experiments. The accuracy of each distance on each dataset is averaged over $10$ runs. The same technique is followed for all other distance families to report accuracy, recall, and precision of the KNN classifier for each distance on each dataset. The average values for each of 54 distances considered in the paper is summarized in Table~\ref{tab:allaverage}, where HasD obtained the highest overall average.

\begin{table}
\centering
	\caption{Average accuracies, recalls, precisions over all datasets for each distance. HasD obtained the highest overall average.}\label{tab:allaverage}
	\begin{tabular}{llll|lllll}
	\hline
	Distance	&	Accuracy	&	Recall	&	Precision	&	Distance	&	Accuracy	&	Recall	&	Precision\\
	\hline
ED		&	0.8001	&	0.6749	&	0.6724	&	PSCSD	&	0.6821	&	0.5528	&	0.5504\\
MD		&	0.8113	&	0.6831	&	0.681	&	DivD		&	0.812	&	0.678	&	0.6768\\
CD		&	0.7708	&	0.656	&	0.6467	&	ClaD		&	0.8227	&	0.6892	&	0.6871\\
LD		&	0.8316	&	0.6964	&	0.6934	&	ASCSD	&	0.6259	&	0.4814	&	0.4861\\
CanD	&	0.8282	&	0.6932	&	0.6916	&	SED		&	0.8001	&	0.6749	&	0.6724\\
SD		&	0.7407	&	0.6141	&	0.6152	&	AD		&	0.8001	&	0.6749	&	0.6724\\
SoD		&	0.7881	&	0.6651	&	0.6587	&	SCSD	&	0.8275	&	0.693	&	0.6909\\
KD		&	0.6657	&	0.5369	&	0.5325	&	MCED	&	0.7973	&	0.6735	&	0.6704\\
MCD		&	0.8113	&	0.6831	&	0.681	&	TopD		&	0.6793	&	0.461	&	0.4879\\
NID		&	0.8113	&	0.6831	&	0.681	&	JSD		&	0.6793	&	0.461	&	0.4879\\
CosD	&	0.803	&	0.6735	&	0.6693	&	JDD		&	0.7482	&	0.5543	&	0.5676\\
ChoD	&	0.7984	&	0.662	&	0.6647	&	JefD		&	0.7951	&	0.6404	&	0.6251\\
JacD		&	0.8024	&	0.6756	&	0.6739	&	KLD		&	0.513	&	0.3456	&	0.3808\\
DicD		&	0.8024	&	0.6756	&	0.6739	&	KDD		&	0.5375	&	0.3863	&	0.4182\\
SCD		&	0.8164	&	0.65		&	0.4813	&	KJD		&	0.6501	&	0.4984	&	0.5222\\
HeD		&	0.8164	&	0.65		&	0.6143	&	TanD		&	0.7496	&	0.5553	&	0.5718\\
BD		&	0.4875	&	0.3283	&	0.4855	&	AvgD	&	0.8084	&	0.6811	&	0.6794\\
MatD	&	0.8164	&	0.65		&	0.5799	&	HamD	&	0.6413	&	0.5407	&	0.5348\\
VWHD	&	0.6174	&	0.4772	&	0.5871	&	MeeD	&	0.415	&	0.1605	&	0.3333\\
VSDF1	&	0.7514	&	0.6043	&	0.5125	&	WIAD	&	0.812	&	0.6815	&	0.6804\\
VSDF2	&	0.6226	&	0.4828	&	0.5621	&	HauD	&	0.5967	&	0.4793	&	0.4871\\
VSDF3	&	0.7084	&	0.5791	&	0.5621	&	CSSD	&	0.4397	&	0.2538	&	0.332\\
MSCSD	&	0.7224	&	0.5876	&	0.3769	&	SPeaD	&	0.8023	&	0.6711	&	0.6685\\
MiSCSD	&	0.6475	&	0.5137	&	0.5621	&	CorD		&	0.803	&	0.6717	&	0.6692\\
PCSD	&	0.6946	&	0.5709	&	0.5696	&	PeaD	&	0.759	&	0.6546	&	0.6395\\
NCSD	&	0.6536	&	0.5144	&	0.5148	&	MotD	&	0.7407	&	0.6141	&	0.6152\\
SquD	&	0.6821	&	0.5528	&	0.5504	&	HasD	&	\textbf{0.8394}	&	\textbf{0.7018}	&	\textbf{0.701}\\
	\hline
	\end{tabular}
\end{table}

\begin{table}
\centering
	\caption{The highest accuracy in each dataset.}\label{tab:highaccuracy}
	\begin{tabular}{lll}
	\hline
Dataset	&	Distance	&	Accuracy \\
\hline
Australian		&	CanD	&	0.8209\\
Balance		&	ChoD,SPeaD,CorD,CosD	&	0.933\\
Banknote		&	CD,DicD,JacD	&	1\\
BCW			&	HasD	&	0.9624\\
Cancer		&	HasD	&	0.9616\\
Diabetes		&	MSCSD	&	0.6897\\
Glass		&	LD,MCED&		0.7111\\
Haberman		&	KLD	&	0.7327\\
Heart		&	HamD	&	0.7714\\
Ionosphere	&	HasD	&	0.9025\\
Liver		&	VSDF1	&	0.6581\\
Monkey1	&	WIAD	&	0.9497\\
Parkinson	&	VSDF1	&	0.9997\\
Phoneme	&	VSDF1	&	0.898\\
QSAR 	&	HasD	&	0.8257\\
Segmen	&	SoD,SD	&	0.9676\\
Sonar	&	MiSCSD	&	0.8771\\
Vehicle	&	LD	&	0.6913\\
Vote		&	ClaD,DivD		&	0.9178\\
Vowel	&	LD	&	0.9771\\
Wholesale		&	AvgD	&	0.8866\\
Wine		&	CanD	&	0.985\\
German	&	ASCSD	&	0.71\\
Iris	&	ChoD,SPeaD,CorD,CosD	&	0.9588\\
Wave 21	&	ED,AD,MCED,SED	&	0.7774\\
Egg	&	ChoD,SPeaD,CorD,CosD	&	0.9725\\
Wave 40	&	ED,AD,MCED,SED	&	0.7587\\
Letter rec.	&	JacD,DicD	&	0.9516\\
\hline
	\end{tabular}
\end{table}

Table~\ref{tab:highaccuracy}, show the highest accuracy on each of the datasets obtained by which of the distances. Based on these results we summarize the following observations.
\begin{itemize}

\item
The distance measures in $L_1$ family outperformed the other distance families in $5$ datasets. LD achieved the highest accuracy in two datasets, namely on Vehicle and Vowel with an average accuracies of $69.13\%$, $97.71\%$ respectively. On the other hand, CanD achieved the highest accuracy in two datasets, Australian and Wine datasets with an average accuracies of $82.09\%$, $98.5\%$ respectively. SD and SoD achieved the highest accuracy on Segmen dataset with an average accuracy of $96.76\%$. 	
Among the $L_p$ Minkowski and $L_1$ distance families, the MD, NID and MCD achieved similar performance with overall accuracies on all datasets; this is due to the similarity between these distances.

\item
In Inner product family, JacD and DicD outperform all other tested distances on Letter rec. dataset with an average accuracy of  $95.16\%$.
Among the $L_p$ Minkowski and $L_1$ distance families, the CD, JacD and DicD outperform the other tested distances on the Banknote dataset with an average accuracy of $100\%$.	

\item
In Squared Chord family, MatD, SCD, and HeD achieved similar performance with overall accuracies on all datasets, this is expected because these distances are very similar.  

\item
In Squared $L_2$ distance measures family, the SquD and PSCSD achieved similar performance with overall accuracy in all datasets, this is due to the similarity between these two distances. The distance measures in this family outperform the other distance families on two datasets, namely, the ASCSD achieved the highest accuracy on the German dataset with an average accuracy of $71\%$. ClaD and DivD achieved the highest accuracy on the Vote dataset with an average accuracy of $91.87\%$.
Among the $L_p$ Minkowski and Squared $L_2$ distance measures family, the ED, SED and AD achieved similar performance in all datasets; this is due to the similarity between these three distances. Also, these distances and MCED outperform the other tested distances in two datasets, Wave21, Wave40 with an average accuracies of $77.74\%$, $75.87\%$ respectively.
Among the $L_1$ distance and Squared $L_2$ families, the MCED and LD achieved the highest accuracy on the Glass dataset with an average accuracy of $71.11\%$.

\item
In Shannon entropy distance measures family, JSD and TopD achieved similar performance with overall accuracies on all datasets, this is due to similarity between both of the distances, as the TopD is twice the JSD. KLD outperforms all the tested distances on Haberman dataset with an average accuracy of $73.27\%$.

\item
The Vicissitude distance measures family outperform the other distance families on $5$ datasets, namely, VSDF1 achieved the highest accuracy in three datasets, Liver, Parkinson and  Phoneme with accuracies of $65.81\%$, $99.97\%$, and $89.8\%$ respectively. MSCSD achieved the highest accuracy on the Diabetes dataset with an average accuracy of $68.79\%$. MiSCSD  also achieved the highest accuracy on Sonar dataset with an average accuracy of $87.71\%$.

\item
The other distance measures family outperforms all other distance families in $7$ datasets.
The WIAD achieved the highest accuracy on Monkey1 dataset with an average accuracy of $94.97\%$. The AvgD also achieved the highest accuracy on the Wholesale dataset with an average accuracy of $88.66\%$. HasD also achieved the highest accuracy in four datasets, namely, Cancer, BCW, Ionosphere and QSAR with an average accuracies of $96.16\%$, $96.24\%$, $90.25\%$, and $82.57\%$ respectively. Finally,  HamD achieved the highest accuracy on the Heart dataset with an average accuracy of  $77.14\%$.
Among the inner product and other distance measures families, the SPeaD, CorD, ChoD and CosD outperform other tested distances in three datasets, namely, Balance, Iris and Egg with an average accuracies of $94.3\%$, $95.88\%$, and $97.25\%$  respectively.
\end{itemize}

\begin{table}
\centering
	\caption{The highest recall in each dataset.}\label{tab:highrecall}
	\begin{tabular}{lll}
	\hline
Dataset	&	Distance	&	Recall\\
\hline
Australian	&	CanD	&	0.8183\\
Balance	&	ChoD,SPeaD,CorD,CosD	&	0.6437\\
Banknote	&	CD,DicD,JacD	&	1\\
BCW		&	HasD	&	0.3833\\
Cancer	&	HasD	&	0.9608\\
Diabetes	&	MSCSD	&	0.4371\\
Glass	&	LD	&	0.5115\\
Haberman	&	DicD,JacD	&	0.3853\\
Heart	&	HamD	&	0.5122\\
Ionosphere	&	LD	&	0.6152\\
Liver		&	VSDF1	&	0.4365\\
Monkey1	&	WIAD	&	0.9498\\
Parkinson	&	VSDF1	&	0.9997\\
Phoneme	&	VSDF1	&	0.8813\\
QSAR 	&	HasD	&	0.8041\\
Segmen	&	SoD,SD	&	0.8467\\
Sonar	&	MiSCSD	&	0.5888\\
Vehicle	&	LD	&	0.5485\\
Vote		&	ClaD,DivD	&	0.9103\\
Vowel	&	LD	&	0.9768\\
Wholesale	&	PCSD	&	0.5816\\
Wine		&	CanD	&	0.7394\\
German	&	ASCSD	&	0.4392\\
Iris		&	ChoD,SPeaD,CorD,CosD	&	0.9592\\
Wave 21	&	ED,AD,MCED,SED	&	0.7771\\
Egg		&	ChoD,SPeaD,CorD,CosD	&	0.9772\\
Wave 40	&	ED,AD,MCED,SED	&	0.7588\\
Letter rec.	&	VSDF2	&	0.9514\\
\hline
	\end{tabular}
\end{table}

Table~\ref{tab:highrecall} shows the highest recalls on each of the datasets obtained by which of the distances. Based on these results we summarize the following observations. 
\begin{itemize}

\item The $L_1$ distance measures family outperform the other distance families in $7$ datasets, for example, CanD achieved the highest recalls in two datasets, Australian and Wine with $81.83\%$ and $73.94\%$ average recalls respectively.  LD also achieved the highest recalls on four datasets, Glass, Ionosphere, Vehicle and Vowel with $51.15\%$, $61.52\%$, $54.85\%$ and $97.68\%$ average recalls respectively. SD and SoD achieved the highest recall on Segmen dataset with $84.67\%$ average recall. 
Among the $L_p$ Minkowski and $L_1$ distance families, the MD, NID and MCD achieved similar performance as expected, due to their similarity.

\item In Inner Product distance measures family, JacD and DicD outperform all other tested distances in Heberman dataset with 38.53\% average recall. 
Among the $L_p$ Minkowski and Inner Product distance measures families, the CD, JacD and DicD also outperform the other tested distances in the Banknote dataset with 100\% average recall.

\item  In Squared chord distance measures  family, MatD, SCD, and HeD achieved similar performance; this is due to their equations similarity as clarified previously.

\item The Squared $L_2$ distance measures family outperform the other distance families on two datasets, namely,  ClaD and DivD outperform the other tested distances on Vote dataset with $91.03\%$ average recall. The PCSD outperforms the other tested distances on Wholesale dataset with $58.16\%$ average recall. ASCSD also outperforms the other tested distances on German dataset with $43.92\%$ average recall.
Among the $L_p$ Minkowski and Squared $L_2$ distance measures families, the ED, SED and AD achieved similar performance in all datasets; this is due to their equations similarity as clarified previously. These distances and MCED distance outperform the other tested distances in two datasets, namely, the Wave21, Wave40 with $77.71\%$ and $75.88\%$ average recalls respectively.

\item In Shannon entropy distance measures family, JSD and TopD distances achieved similar performance as expected, due to their similarity.

\item The Vicissitude distance measures family outperform the other distance families in six datasets. The VSDF1 achieved the highest recall on three datasets, Liver, Parkinson and Phoneme datasets with $43.65\%$, $99.97\%$, $88.13\%$ average recalls respectively. MSCSD achieved the highest recall on Diabetes dataset with $43.71\%$ average recall. MiSCSD also achieved the highest recall on Sonar dataset with $58.88\%$ average recall. The VSDF2 achieved the highest recall on Letter rec. dataset with $95.14\%$ average recall.

\item The other distance measures family outperforms the all other distance families in $5$ datasets. Particularly, HamD achieved the highest recall on the Heart dataset with $51.22\%$ average recall. The WIAD also achieved the highest average recall on the Monkey1 dataset with $94.98\%$ average recall. HasD also has achieved the highest average recall on three datasets, namely, Cancer, BCW and QSAR with $96.08\%$, $38.33\%$, and $80.41\%$ average recalls respectively.
Among the inner product and other distance measures families, the SPeaD, CorD, ChoD and CosD outperform the other tested distances in three datasets, namely, Balance, Iris and Egg with $64.37\%$, $95.92\%$, and $97.72\%$ average recalls respectively.

\end{itemize}
 
\begin{table}
\centering
	\caption{The highest precision in each dataset.}\label{tab:highprecision}
	\begin{tabular}{lll}
	\hline
Dataset	&	Distance	& Precision\\
\hline
Australian	&	CanD	&	0.8188\\
Balance	&	ChoD,SPeaD,CorD,CosD	&	0.6895\\
Banknote	&	CD,DicD,JacD	&	1\\
BCW		&	HasD	&	0.3835\\
Cancer	&	HasD	&	0.9562\\
Diabetes	&	ASCSD	&	0.4401\\
Glass	&	LD	&	0.5115\\
Haberman	&	SPeaD,CorD,CosD	&	0.3887\\
Heart	&	HamD	&	0.5112\\
Ionosphere	&	HasD	&	0.5812\\
Liver		&	VSDF1	&	0.4324\\
Monkey1	&	WIAD	&	0.95\\
Parkinson	&	VSDF1	&	0.9997\\
Phoneme	&	VSDF1	&	0.8723\\
QSAR 	&	HasD	&	0.8154\\
Segmen	&	SoD,SD	&	0.8466\\
Sonar	&	MiSCSD	&	0.5799\\
Vehicle	&	LD	&	0.5537\\
Vote		&	ClaD,DivD		&	0.9211\\
Vowel	&	LD		&	0.9787\\
Wholesale	&	DicD,JacD	&	0.5853\\
Wine		&	CanD	&	0.7408\\
German	&	ASCSD	&	0.4343\\
Iris		&	ChoD,SPeaD,CorD,CosD	&	0.9585\\
Wave 21	&	ED,AD,MCED,SED	&	0.7775\\
Egg		&	ChoD,SPeaD,CorD,CosD	&	0.9722\\
Wave 40	&	ED,AD,MCED,SED	&	0.759\\
Letter rec.	&	ED,AD,SED	&	0.9557\\
\hline
	\end{tabular}
\end{table}

Table~\ref{tab:highprecision} show the highest precisions on each of the datasets obtained by which of the distances. Based on these results we summarize the following observations. 
\begin{itemize}

\item The distance measures in $L_1$ family outperformed the other distance families in $5$ datasets. CanD achieved the highest precision on two datasets, namely, Australian and  Wine with $81.88\%$, $74.08\%$ average precisions respectively. SD and SoD achieved the highest precision on the Segmen dataset with $84.66\%$ average precision. In addition, LD  achieved the highest precision on three datasets, namely, Glass, Vehicle and Vowel, with $51.15\%$, $55.37\%$, and $97.87\%$ average precisions respectively. 
Among the $L_p$ Minkowski and $L_1$ distance families, the MD, NID and MCD achieved similar performance in all datasets; this is due to their equations similarity as clarified previously.

\item Inner Product family outperform other distance families in two datasets. Also, JacD and DicD outperform the other tested measures on Wholesale dataset with 58.53\% average precision.
Among the $L_p$ Minkowski and $L_1$ distance families, that the  CD, JacD and DicD on the other tested distances on the Banknote dataset with $100\%$ average precision.

\item In Squared chord distance measures family, MatD, SCD, and HeD achieved similar performance with overall precision results in all datasets; this is due to their equations similarity as clarified previously.

\item  In Squared $L_2$ distance measures family, SCSD and PSCSD achieved similar performance; this is due to their equations similarity as clarified previously. The distance measures in this family outperform the other distance families on three datasets, namely,  ASCSD  achieved the highest average precisions on two datasets, Diabetes and German with $44.01\%$, and $43.43\%$ average precisions respectively. ClaD and DicD also achieved the highest precision on the Vote dataset with $92.11\%$ average precision.
Among the $L_p$ Minkowski and  Squared $L_2$ distance measures families, the ED, SED and AD achieved similar performance as expected, due to their similarity. These distances and MCED outperform  the other tested measures  in two  datasets, namely, Wave 21, Wave 40 with $77.75\%$, $75.9\%$ average precisions respectively. Also, ED, SED and AD outperform the other tested measures  on  the Letter rec. with $95.57\%$ average precision.      

\item In Shannon entropy distance measures family, JSD and TopD achieved similar performance with overall precision in all datasets, due to their equations similarity as clarified earlier.

\item The Vicissitude distance measures family outperform other distance families on four datasets. The VSDF1 achieved the highest average precisions on three datasets, Liver, Parkinson and Phoneme with $43.24\%$, $99.97\%$, $87.23\%$ average precisions respectively. MiSCSD also achieved the highest precision on the Sonar dataset with $57.99\%$ average precision.

\item The other distance measures family outperforms all the other distance families in $6$ datasets. In particular, HamD achieved the highest precision on the Heart dataset with $51.12\%$ average precision. Also, WIAD achieved the highest precision on the Monkey1 dataset with $95\%$ average precision. Moreover, HasD yield the highest precision in four datasets, namely, Cancer, BCW, Ionosphere and QSR, with $38.35\%$, $95.62\%$, $58.12\%$, and $81.54\%$ average precisions respectively.
Among the inner product and other distance measures families, the SPeaD, CorD, ChoD and CosD outperform the other tested distances in three datasets, namely, Balance, Iris and Egg, with $68.95\%$, $95.85\%$, and $97.22\%$ average precisions respectively. Also, CosD, SPeaD, CorD achieved the highest precision on Heberman dataset with $38.87\%$ average precision.

\end{itemize}

\begin{table}
\centering
	\caption{The top $10$ distances in terms of average accuracy, recall, and precision based performance on noise-free datasets.}\label{tab:top10}
	\begin{tabular}{lllllllll}
	\hline
	&	Accuracy				&	&	&	Recall	&		&	&	Precision	\\
Rank	&	Distance	&	Average		&	Rank &	Distance	&	Average		&	Rank	&	Distance	&	Average\\
	\hline
1	&	HasD	&	0.8394	&	1	&	HasD	&	0.7018	&	1&	HasD	&	0.701\\
2	&	LD		&	0.8316	&	2	&	LD		&	0.6964	&	2&	LD		&	0.6934\\
3	&	CanD	&	0.8282	&	3	&	CanD	&	0.6932	&	3&	CanD	&	0.6916\\
4	&	SCSD	&	0.8275	&	4	&	SCSD	&	0.693	&	4&	SCSD	&	0.6909\\
5	&	ClaD		&	0.8227	&	5	&	ClaD		&	0.6892	&	5&	ClaD		&	0.6871\\
6	&	DivD		&	0.812	&	6	&	MD		&	0.6831	&	6&	MD		&	0.681\\
6	&	WIAD	&	0.812	&	7	&	WIAD	&	0.6815	&	7&	WIAD	&	0.6804\\
7	&	MD		&	0.8113	&	8	&	AvgD	&	0.6811	&	8&	DivD		&	0.6768\\
8	&	AvgD	&	0.8084	&	9	&	DivD		&	0.678	&	9&	DicD		&	0.6739\\
9	&	CosD	&	0.803	&	10	&	DicD		&	0.6756	&	10&	ED		&	0.6724\\
9	&	CorD		&	0.803							\\	
10	&	DicD		&	0.8024								\\
	\hline
	\end{tabular}
\end{table}

Table~\ref{tab:top10} shows the top $10$ distances in respect to the overall average accuracy, recall and precision over all datasets. HasD outperforms all other tested distances in all performance measures, followed by LD, CanD and SCSD. Moreover, a closer look at the data of the average as well as highest accuracies, precisions, recalls, we find that the HasD outperform all distance measures on $4$ datasets, namely, Cancer, BCW, Ionosphere and QSAR, this is true for accuracy, precision and recall, and it is the only distance metric that won at least $4$ datasets in this noise-free experiment set.   
Note that the performance of the following five group members, (1) MCD, MD, NID (2) AD, ED SED, (3) TopD, JSD, (4) SquD, PSCSD, and (5) MatD, SCD, and HeD are the same within themselves due to their close similarity in defining the corresponding distances.
                                                       
\begin{table}
\centering
	\caption{The p-values of the Wilcoxon test for the results of Hassanat distance with each of other top distances over the datasets used. The p-values that were less than the significance level ($0.05$) are highlighted in \textbf{boldface}.}\label{tab:pvalues}
	\begin{tabular}{llll}
	\hline
Distance	&	Accuracy	&	Recall	&	Precision\\
	\hline
ED		&	\textbf{0.0418}	&	0.0582	&	\textbf{0.0446}\\
MD		&	0.1469	&	0.1492	&	0.1446\\
CanD	&	\textbf{0.0017}	&	\textbf{0.008}	&	\textbf{0.0034}\\
LD		&	0.0594	&	0.121	&	\textbf{0.0427}\\
CosD	&	\textbf{0.0048}	&	\textbf{0.0066}	&	\textbf{0.0064}\\
DicD		&	0.0901	&	0.0934	&	0.0778\\
ClaD		&	\textbf{0.0089}	&	\textbf{0.0197}	&	\textbf{0.0129}\\
SCSD	&	\textbf{0.0329}	&	0.0735	&	0.0708\\
WIAD	&	\textbf{0.0183}	&	\textbf{0.0281}	&	\textbf{0.0207}\\
CorD		&	\textbf{0.0048}	&	\textbf{0.0066}	&	\textbf{0.0064}\\
AvgD	&	0.1357	&	0.1314	&	0.102\\
DivD		&	\textbf{0.0084}	&	\textbf{0.0188}	&	\textbf{0.017}\\
      	\hline
	\end{tabular}
\end{table}            

We attribute the success of Hassanat distance in this experimental part to its characteristics discussed in Section~\ref{ssec:dist} (See distance equation in 8.13, Figure~\ref{fig:hass}), where each dimension in the tested vectors contributes maximally 1 to the final distance, this lowers and neutralizes  the effects of outliers in different datasets. 
To further analyze the performance of Hassanat distance comparing with other top distances we used the Wilcoxon's rank-sum test~\citep{Wilcoxon1945}. This is a non-parametric pairwise test that aims to detect significant differences between two sample means, to judge if the null hypothesis is true or not. Null hypothesis is a hypothesis used in statistics that assumes there is no significant difference between different results or observations. This test was conducted between Hassanat distance and with each of the other top distances (see Table~\ref{tab:top10}) over the tested datasets. Therefore, our null hypothesis is: ``there is no significant difference between the performance of Hassanat distance and the compared distance over all the datasets used". 
According to the Wilcoxon test, if the result of the test showed that the P-value is less than the significance level ($0.05$) then we reject the null hypothesis, and conclude that there is a significant difference between the tested samples; otherwise we cannot conclude anything about the significant difference~\citep{Derrac2011}.

The accuracies, recalls and precisions of Hassanat distance over all the datasets used in this experiment set were compared to those of each of the top $10$ distance measures, with the corresponding p-values are given in Table~\ref{tab:pvalues}.  The p-values that were less than the significance level ($0.05$) are highlighted in bold. As can be seen from Table~\ref{tab:pvalues}, the p-values of accuracy results is less than the significance level ($0.05$) eight times, here we can reject the null hypothesis and conclude that there is a significant difference in the performance of Hassanat distance compared to ED, CanD, CosD, ClaD, SCSD, WIAD, CorD and DivD, and since the average performance of Hassanat distance was better than all of these distance measures from the previous tables, we can conclude that the accuracy yielded by Hassanat distance is better than that of most of the distance measures tested. Similar analysis applies for the recall, and precision columns comparing Hassanat results to the other distances.
\subsection{With noise}\label{ssec:with}

These next experiments aim to identify the impact of noisy data on the performance of KNN classifier regarding accuracy, recall and precision using different distance measures.   Accordingly, nine different levels of noise were added into each dataset using Algorithm~\ref{alg2}. For simplicity, this set of experiments conducted using only the top $10$ distances shown in Table~\ref{tab:top10} that are obtained based on the noise-free datasets. 

\begin{figure}
\centering
	\includegraphics[width= 10cm]{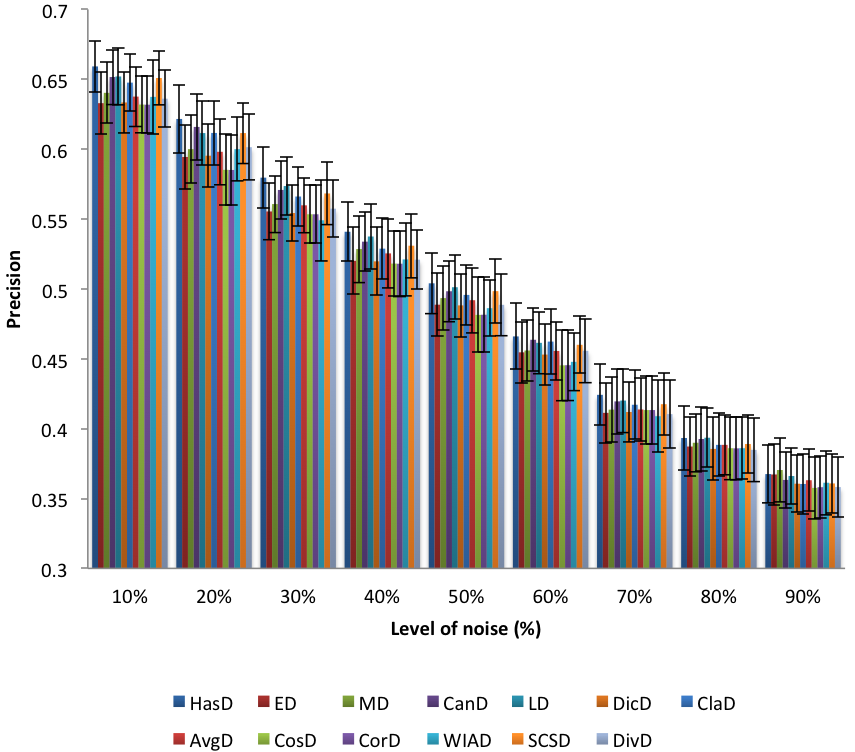}
	\caption{The overall average accuracies and standard deviations of KNN classifier using top 10 distance measures with different levels of noise.}\label{fig:accuracy_noisy}
\end{figure}

\begin{figure}
\centering
	\includegraphics[width= 10cm]{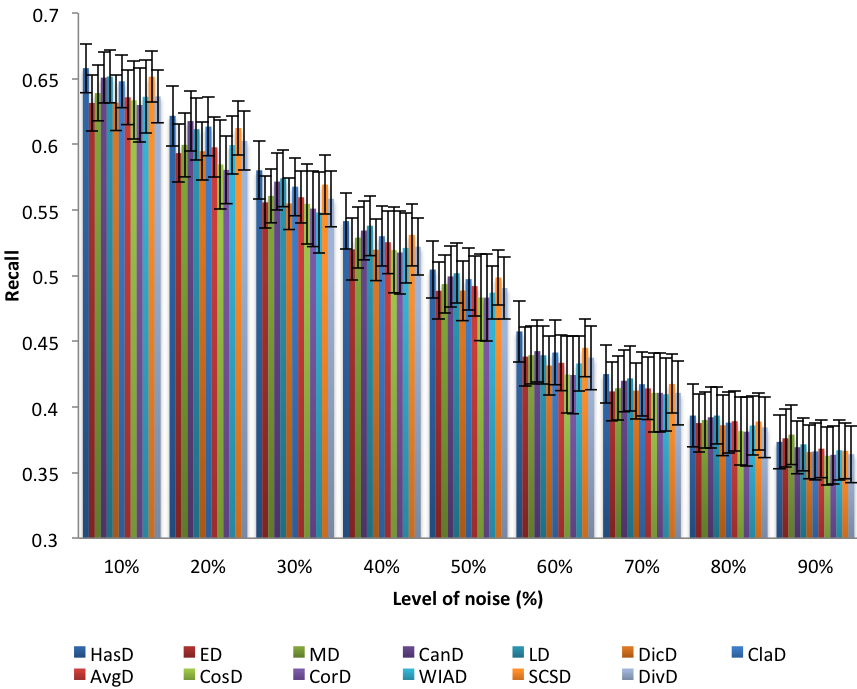}
	\caption{The overall average recalls and standard deviations of KNN classifier using top 10 distance measures with different levels of noise.}\label{fig:recall_noisy}
\end{figure}

\begin{figure}
\centering
	\includegraphics[width= 10cm]{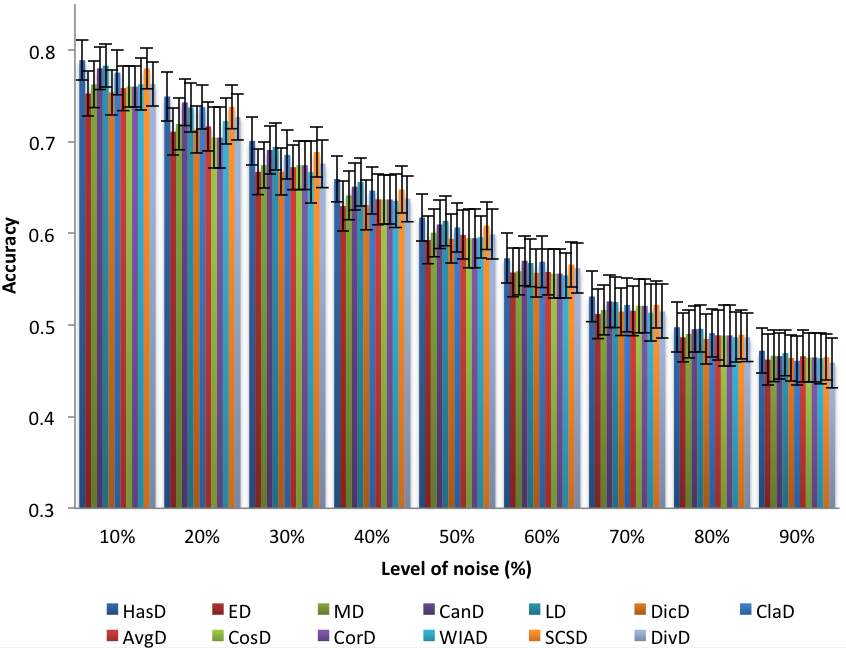}
	\caption{The overall average precisions and standard deviations of KNN classifier using top 10 distance measures with different levels of noise.}\label{fig:precision_noisy}
\end{figure}

\begin{table}
\centering
	\caption{Ranking of distances in descending order based on the accuracy results at each noise level.}\label{tab:rank_accuracy}
	\begin{tabular}{llllllllll}
	\hline
	Rank	&10\%	&20\%	&30\%	&40\%	&50\%	&60\%	&70\%	&80\%	&90\%\\
	\hline
	1	&HasD	&HasD	&HasD	&HasD	&HasD	&HasD	&HasD	&HasD	&HasD\\
	\hline
	2	&LD		&CanD	&LD		&LD		&LD		&CanD	&CanD	&LD		&LD\\
	\hline
	\multirow{2}{*}{3}	&CanD	&SCSD	&CanD	&CanD	&CanD	&ClaD	&LD		&CanD	&MD\\
		&SCSD	&&&&&&&\\
	\hline
	4	&ClaD	&ClaD	&SCSD	&SCSD	&SCSD	&LD		&SCSD	&ClaD	&CanD\\
	\hline
	5	&DivD	&LD		&ClaD	&ClaD	&ClaD	&SCSD	&ClaD	&MD		&AvgD\\
	\hline
	\multirow{2}{*}{6}	&WIAD	&DivD	&DivD	&MD		&MD		&DivD	&CosD	&SCSD	&SCSD\\
		&		&		&		&		&		&		&CorD	&		&	\\
	\hline
	\multirow{3}{*}{7}	&MD		&WIAD	&MD		&DivD	&DivD	&MD		&MD		&AvgD	&CorD\\
					&		&		&CosD\\
					&		&		&CorD	\\
	\hline					
	8	&CD		&MD		&AvgD	&AvgD	&AvgD	&AvgD	&AvgD	&CorD	&CosD\\
	\hline
	\multirow{2}{*}{9}	&CorD	&AvgD	&DicD	&CosD	&WIAD	&ED		&DivD	&CosD	&DicD\\
		&		&		&		&CorD	\\
	\hline
	\multirow{3}{*}{10}	&AvgD	&DicD	&ED		&WIAD	&CoD	&DicD	&DicD	&DivD	&WIAD\\
					&		&		&		&		&		&		&&WIAD\\
					&		&		&		&		&		&		&&ED	\\
	\hline																
	11	&DicD	&ED		&WIAD	&DicD	&CorD	&CorD	&WIAD	&DicD	&ED\\
		&		&		&		&		&		&CosD	\\
	\hline	
	12	&ED		&CosD	&-		&ED		&DicD	&WIAD	&ED		&-		&ClaD\\
	\hline
	13	&-		&CorD	&-		&-		&ED		&-		&-		&-		&DivD\\
	\hline
	\end{tabular}
\end{table}

\begin{table}
\centering
	\caption{Ranking of distances in descending order based on the recall results at each noise level.}\label{tab:rank_recall}
	\begin{tabular}{llllllllll}
	\hline
	Rank	&10\%	&20\%	&30\%	&40\%	&50\%	&60\%	&70\%	&80\%	&90\%\\
	\hline
	1			&HasD	&HasD	&HasD	&HasD	&HasD	&HasD	&HasD	&LD HasD	&MD\\
	\hline
	2			&LD		&CanD	&LD		&LD		&LD		&SCSD	&LD		&CanD	&ED\\
	\hline
	3			&SCSD	&ClaD	&CanD	&CanD	&CanD	&CanD	&CanD	&MD		&HasD\\
	\hline
	4			&CanD	SCSD	&SCSD	&SCSD	&SCSD	&ClaD	&SCSD	&AvgD	&LD\\
	\hline
	5			&ClaD	&LD		&ClaD	&ClaD	&ClaD	&MD		&ClaD	&SCSD	&CanD\\
	\hline
	6			&MD		&DivD	&MD		&MD		&MD		&LD		&MD		&ClaD	&AvgD\\
	\hline
	7			&DivD	&MD		&AvgD	&AvgD	&AvgD	&ED		&AvgD	&ED		&WIAD\\
	\hline
	8			&WIOD	&WIOD	&DivD	&DivD	&DivD	&DivD	&DicD	&DicD	&SCSD\\
	\hline
	9			&AvgD	&AvgD	&ED		&WIAD	&DicD	&AvgD	&ED		&WIAD	&ClaD\\
	\hline
	\multirow{3}{*}{10}		&CosD	&DicD	&DicD	&ED		&ED		&WIAD	&CosD	&DivD	&DicD\\
						&		&		&		&		&		&		&CorD\\
						&		&		&		&		&		&		&DivD	\\
	\hline
	11			&DicD	&ED		&CosD	&DicD	&WIAD	&DicD	&WIAD	&CosD	&DivD\\
	\hline
	12			&ED		&CosD	CorD		&CosD	&CoD	&CosD	&-		&CorD	&CorD\\
	\hline
	13	&CorD	&CorD	&WIOD	&CorD	&CorD	&CorD	&-	&-	&CosD\\
	\hline
	\end{tabular}
\end{table}	

\begin{table}
\centering
	\caption{Ranking of distances in descending order based on the precision results at each noise level.}\label{tab:rank_precision}
	\begin{tabular}{llllllllll}
	\hline
	Rank	&10\%	&20\%	&30\%	&40\%	&50\%	&60\%	&70\%	&80\%	&90\%\\
	\hline
	1	&HasD	&HasD	&HasD	&HasD	&HasD	&HasD	&HasD	&LD		&MD\\
	\hline
	2	&LD		&CanD	&LD		&LD		&LD		&CanD	&LD		&HasD	&HasD\\
	\hline
	3	&CanD	&ClaD	&CanD	&CanD	&SCSD	&ClaD	&CanD	&CanD	&ED\\
	\hline
	\multirow{2}{*}{4}	&SCSD	&LD		&SCSD	&SCSD	&CanD	&LD		&SCSD	&MD		&LD\\
					&		&SCSD	\\
	\hline
	5	&ClaD	&DivD	&ClaD	&ClaD	&ClaD	&SCSD	&ClaD	&SCSD	&CanD\\
	\hline
	\multirow{2}{*}{6}	&MD		&MD		&MD		&MD		&MD		&MD		&AvgD	&ClaD	&AvgD\\
		&			&		&WIAD	\\
	\hline
	7	&AvgD	&AvgD	&AvgD	&AvgD	&AvgD	&DivD	&MD		&AvgD	&WIAD\\
	\hline
	8	&WIAD	&DicD	&DivD	&WIAD	&ED		&AvgD	&CosD	&ED		&SCSD\\
	&		&		&		&		&		&		&CorD	\\
	\hline
	9	&DivD	&ED		&ED		&DivD	&DivD	&ED		&DicD	&CosD	&DicD\\
		&		&		&		&		&		&			&	&WIAD\\
	\hline
	10	&DicD	&CosD	&DicD	&ED		&DicD	&DicD	&ED		&CorD	&ClaD\\
	\hline
	11	&ED		&CorD	&CosD	&DicD	&WIAD	&WIAD	&DivD	&DicD	&DivD\\
		&		&		&CorD	&	\\
	\hline
	12	&CosD	&-		&WIAD	&CosD	&CorD	&CorD	&WIAD	&DivD	&CorD\\
	\hline
	13	&CorD	&-		&-		&CorD	&CosD	&CosD	&-			&-	&CosD\\
	\hline
	\end{tabular}
\end{table}

Figure~\ref{fig:accuracy_noisy} shows the experimental results of KNN classifier that clarify  the impact of noise on  the accuracy performance measure using the top $10$ distances. X-axis denotes the noise level and Y-axis represents the classification accuracy. Each column at each noise level represents the overall  average accuracy for each distance on all datasets used. Error bars represent the average of  standard deviation values for each distance on all datasets.
Figure~\ref{fig:recall_noisy} shows the recall results of KNN classifier that clarify  the impact of noise on the performance using the top $10$ distance measures.  
Figure~\ref{fig:precision_noisy} shows the precision results of KNN classifier that clarify  the impact of noise on the performance using the top $10$ distance measures.
As can be seen from Figures~\ref{fig:accuracy_noisy}, \ref{fig:recall_noisy} and \ref{fig:precision_noisy} the performance (measured by accuracy, recall, and precision respectively) of the KNN degraded only about $20\%$ while the noise level reaches $90\%$, this is true for all the distances used. This means that the KNN classifier using any of the top $10$ distances tolerate noise to a certain degree. 
Moreover, some distances are less affected by the added noise comparing to other distances.  Therefore, we ordered the distances according to their overall average accuracy, recall and precision results for each level of noise. The distance with highest performance is ranked in the first position, while the distance with the lowest performance is ranked in the last position of the order. Tables~\ref{tab:rank_accuracy}, \ref{tab:rank_recall} and \ref{tab:rank_precision} show this ranking structure in terms of accuracy, precision, and recall under each noise level from low $10\%$ to high $90\%$. The empty cells occur because of sharing same rank by more than one distance.
The following points summarize the observations in terms of accuracy, precision, and recall values: 
\begin{itemize}

\item 
According to the average precision results, the highest precision was obtained by HasD which achieved the first rank in the majority of noise levels. This distance succeeds to be in the first rank at noise levels $10\%$ up to $70\%$. However, at a level $80\%$, LD outperformed HasD. Also, MD outperformed on the HasD at a noise level $90\%$. 

\item
LD achieved the second rank at noise levels $10\%$, $30\%$, $40\%$, $50\%$, and $70\%$.The CanD achieved the second rank at noise levels 20\% and 60\%. Moreover, this distance achieved the third rank in the rest noise levels except at noise levels $50\%$ and $90\%$. The SCSD achieved the fourth rank at noise levels $10\%$, $30\%$, $40\%$, and $70\%$ and the third rank at a level of noise $50\%$. This distance was equal with the LD at a noise level $20\%$.  The ClarD achieved the third rank at noise levels $20\%$, and $60\%$. 

\item 
The rest of distances achieved the middle and the last ranks in different orders at each level of noise. The cosine distance at level $80\%$ was equal to the WIAD in the result.  This distance was also equal with the CorD at levels $30\%$ and $70\%$. These two distances performed  the worst (lowest precision) in most noise levels.

\end{itemize}

Based on results in Tables~\ref{tab:rank_accuracy}, \ref{tab:rank_recall} and \ref{tab:rank_precision}, we observe that the ranking of distances in terms of accuracy, recall and precision  without the presence of noise is different with their  ranking when adding the first level of noise 10\% and  it become variants significantly when we increased the level of noise progressively. This means that the distances are affected by noise. However, the crucial question is: which one is the distances is least affected by noise? From the above results we conclude that HasD is the least affected one, followed by LD, CanD and SCSD. The good performance of the KNN achieved by these distances  might be contributed by their good characteristics, Table~\ref{table15} shows these characteristics of the top-10 distances in our analysis here. All of these top-10 distances are symmetric, and we further provide input, output ranges and the number of operations.

\subsection{Precise evaluation of the effects of noise}\label{ssec:precise}

In order to justify why some distances are affected either less or more by noise, the following toy Examples~\ref{ex1} and \ref{ex2} are designed. These illustrate the effect of noise on the final decision of the KNN classifier using Hassanat (HasD) and the standard Euclidean (ED) distances. 
In both examples, we assume that we have two training vectors (v1 and v2) having three attributes for each, in addition to one test vector (v3). As usual, we calculate the distances between v3 and both v1 and v2 using both of Euclidean and Hassanat distances.
\begin{example}\label{ex1}
This example shows the KNN classification using two different distances on clean data (without noise). We find the Distance to test vector (v3) according to ED and HasD.

\begin{tabular}{l|lllll|ll}
\hline
	&	X1	&	X2	&	X3	&	X4	&	Class	&	Dist($\cdot$,V3)\\
	&		&		&		&		&			&	ED	&	HasD\\
	\hline
V1	&	3	&	4	&	5	&	3	&	2		&	2	&	0.87\\
V2	&	1	&	3	&	4	&	2	&	1		&	1	&	0.33\\
V3	&	2	&	3	&	4	&	2	&	?		&		&	\\
\hline
\end{tabular}

\end{example}
As can be seen, assuming that we use k = 1, based on the 1-NN approach, and using both distances,  the test vector is assigned to class 1, both results are reasonable, because V3 is almost the same as V2 (class =1) except for the 1st feature, which differs only by 1.
\begin{example}\label{ex2}
This example shows the same feature vectors as in Example~\ref{ex1}, but after corrupting one of the features with an added noise. That is, we make the same previous calculations using noisy data instead of the clean data; the first attribute in V2 is corrupted by an added noise of (4, i.e. X1 = 5).

\begin{tabular}{l|lllll|ll}
\hline
	&	X1	&	X2	&	X3	&	X4	&	Class	&	Dist($\cdot$,V3)\\
	&		&		&		&		&			&	ED	&	HasD\\
	\hline
V1	&	3	&	4	&	5	&	3	&	2		&	2	&	0.87\\
V2	&	5	&	3	&	4	&	2	&	1		&	3	&	0.5\\
V3	&	2	&	3	&	4	&	2	&	?		&		&	\\
\hline
\end{tabular}

\end{example}

Based on the minimum distance approach, using Euclidian distance, the test vector is assigned to class 2 instead of 1. However, the test vector is assigned to class 1 using the Hassanat distance, this makes the distance more accurate with the existence of noise. 
Although simple, these examples showed that the Euclidean distance was affected by noise and consequently affected the KNN classification ability. Although the performance of the KNN classifier is decreased as the noise increased (as shown by the extensive experiments with various datasets), we find that some distances are less affected by noise than other distances. For example, when using ED any change in any attribute contributes highly to the final distance, even if both vectors were similar but in one feature there was noise, the distance (in such a case) becomes unpredictable. In contrast, with the Hassanat distance we found that the distance between both consecutive  attributes are bounded in the range $[0,1]$, thus, regardless of the value of the added noise, each feature will contributes up to 1 maximally to the final distance, and not proportional to the value of the added noise. Therefore the impact of noise on the final classification is mitigated.

Note that the above experiments had the value of K to be $1$. In fact, the choice of distance metric might affect the optimal K as well. A $K = 1$ choice is more sensitive to noise than larger K values, because an unrelated noisy example might be the nearest to a test example. Therefore, a valid action with noisy data would be to choose a larger K; it would be of interest to see which distance measure handles this aspect best. We remark that choosing the optimal K is out of the scope of this review, and we refer to~\citep{Hassanat2014b} and~\citep{Alkasassbeh2015}. However, we have repeated the experiments on all datasets using $K =3$ and $k=\sqrt{n}$, as done by~\citep{Hassanat2014b}, where $n$ is the number of examples in the training dataset, with $50\%$ noise. This is done using the top-10 distances. As can be seen from Table~\ref{table16}, the average accuracy of most of the top-10 distances have been slightly improved compared to Figure~\ref{fig:accuracy_noisy} with noisy data, and Table~\ref{tab:top10} without noise, this is due to the larger number of K used, some exceptions, including the WIAD distance, which seems to be negatively affected by increasing the number of neighbors (K).

In the previous experiments, the noisy data have been used with the top-10 distances only. However, it would be interesting to see if any of the other measures would handle noise better than these particular top-10 measures. Table~\ref{table17} shows the average accuracy of all distances over the first 14 datasets, using $K = 1$ with and without noise.
As can be seen from Table~\ref{table17}, some of the top-10 distances still remain ranked the highest even with the existence of noise compared to all other distances, this includes HasD, LD, DivD, CanD, ClaD and SCSD. However, interestingly, some of the other distances (which were ranked low when using data without noise), have shown less vulnerability to noise, these include SquD, PSCSD, MSCSD, SCD, MatD and HeD. According to the extensive experiments conducted for the purpose of this review, and regardless of the type of the experiments, in general, the non-convex distance HasD is the best distance to be used with the KNN classifier, with other distances such as LD, DivD, CanD, ClaD, and SCSD performing close to best.

\begin{table}[!ht]
  \renewcommand{\arraystretch}{1.3}
  \caption{Some characteristics of the top-10 distances ($n$ is the number of features).}  \label{table15} 
 
  \small
  \begin{center}
   {
    \begin {tabular}{|l|c|c|c|c|c|} 
    \hline
    
 \textbf{distance}	&\textbf{Input} \textbf{range}	&\textbf{Output} \textbf{range}	&\textbf{Symmetric} 	&\textbf{Metric}	&\textbf{\#Operations}\\
\hline
\hline
HasD	&$(-\infty,+\infty)$	&[0,n]	&yes	&yes	&6n (positives)9n (negatives) \\
\hline
ED	    &$(-\infty,+\infty)$	&$[0, +\infty)$	&yes	&yes	&1+3n\\
\hline
MD	    &$(-\infty,+\infty)$	&$[0, +\infty)$	&yes	&yes	&3n\\
\hline
CanD	&$(-\infty,+\infty)$	&$[0, n]$	&yes	&yes	&7n\\
\hline
LD	    &$(-\infty,+\infty)$	&$[0, +\infty)$	&yes	&yes	&5n\\
\hline
CosD	&$[0, +\infty)$	&$[0, 1]$	&yes	&No	   &5+6n\\
\hline
DicD	&$[0, +\infty)$	&$[0, 1]$	&yes	&No	    &4+6n\\
\hline
ClaD	&$[0, +\infty)$	&$[0, n]$	&yes	&yes	&1+6n\\
\hline
SCSD	&$(-\infty,+\infty)$	&$[0, +\infty)$	&yes	&yes	&6n\\
\hline
WIAD	&$(-\infty,+\infty)$	&$[0, 1]$ &yes	&yes	&1+7n\\
\hline
CorD	&$[0, +\infty)$	&$[0, 1]$	&yes	&No	&8+10n\\
\hline
AvgD	&$(-\infty,+\infty)$	&$[0, +\infty)$ &yes	&yes	&2+6n\\
\hline
DivD	&$(-\infty,+\infty)$	&$[0, +\infty)$	&yes	&yes	&1+6n\\
\hline

    \hline 
    
   \end{tabular}

  }
 \end{center}
\end{table}


\begin{table}[!ht]
  \renewcommand{\arraystretch}{1.3}
  \caption{The average accuracy of the top-10 distances over all datasets, using $K=3$ and $K=\sqrt{n}$ (n is the number of features), with and without noise.}\label{table16} 
  \small
  \begin{center}
   {
    \begin {tabular}{|l|c|c|c|c|} 
    \hline

   \multirow{2}{*}{\textbf{Distance}} &\multicolumn{2}{c|}{\textbf{With 50\% noise}} &\multicolumn{2}{c|}{\textbf{Without noise}}\\
   \cline{2-5}
       &$K=3$ 	&$K=\sqrt{n}$	&$K=3$	&$K=\sqrt{n}$\\    
    \hline
    \hline
HasD	&0.6302	&0.6314	&0.8497	&0.833721\\
\hline
LD	    &0.6283	&0.6237	&0.8427	&0.827561\\
\hline
CanD	&0.6247	&0.6231	&0.8384	&0.825171\\
\hline
ClaD	&0.6171	&0.6136	&0.8283	&0.8098\\
\hline
SCSD	&0.6146	&0.6087	&0.8332	&0.8045\\
\hline
MD	    &0.6124	&0.6089	&0.8152	&0.79705\\
\hline
DivD	&0.6114	&0.6049	&0.8183	&0.792768\\
\hline
CosD	&0.6086	&0.6021	&0.8085	&0.791625\\
\hline
CorD	&0.6085	&0.6020	&0.8085	&0.786696\\
\hline
AvgD	&0.6079	&0.6081	&0.8119	&0.792768\\
\hline
ED	    &0.6016	&0.6011	&0.8031	&0.781639\\
\hline
DicD	&0.5998	&0.6016	&0.8021	&0.778054\\
\hline
WIAD	&0.5680	&0.5989	&0.7935	&0.788361\\

\hline

   \end{tabular}
  }
 \end{center}
\end{table}


\begin{table}[!htbp]
  \renewcommand{\arraystretch}{1.3}
  \caption{The average accuracy of all distances over the first 14 datasets, using $K=1$ with and without noise.}\label{table17} 
  \label{table1}
  \begin{center}
   {
    \scriptsize
    \begin {tabular}{|l|c|c|} 
    \hline
    
 \textbf{Distance}	&\textbf{With 50\% noise} 	&\textbf{Without noise} \\
\hline
\hline

\hline
HasD	&0.6331  &0.8108\\
\hline
LD	    &0.6302	&0.7975\\
\hline
DivD	&0.6284	&0.8068\\
\hline
CanD	&0.6271	&0.8053\\
\hline
ClaD	&0.6245	&0.7999\\
\hline
SquD	&0.6227	&0.7971\\
\hline
PSCSD	&0.6227	&0.7971\\
\hline
SCSD	&0.6227	&0.7971\\
\hline
MSCSD	&0.6223	&0.8004\\
\hline
SCD	    &0.6208	&0.7989\\
\hline
MatD	&0.6208	&0.7989\\
\hline
HeD	    &0.6208	&0.7989\\
\hline
VSDF3	&0.6179	&0.7891\\
\hline
WIAD	&0.6144	&0.7877\\
\hline
CorD	&0.6119	&0.7635\\
\hline
SPeaD	&0.6119	&0.7635\\
\hline
CosD	&0.6118	&0.7636\\
\hline
NCSD	&0.6108	&0.7674\\
\hline
ChoD	&0.6102	&0.755\\
\hline
JeD	    &0.6071	&0.7772\\
\hline
PCSD	&0.6043	&0.7813\\
\hline
KJD	    &0.6038	&0.7465\\
\hline
PeaD	&0.6034	&0.7066\\
\hline
VSDF2	&0.6032	&0.7473\\
\hline
MD	&0.6029	&0.7565\\
\hline
NID	&0.6029	&0.7565\\
\hline
MCD	&0.6029	&0.7565\\
\hline
SD	&0.6024	&0.7557\\
\hline
SoD	&0.6024	&0.7557\\
\hline
MotD	&0.6024	&0.7557\\
\hline
ASCSD	&0.6016	&0.7458\\
\hline
VSDF1	&0.6012	&0.7427\\
\hline
AvgD	&0.5995	&0.7523\\
\hline
TanD	&0.5986	&0.7421\\
\hline
JDD	    &0.5955	&0.741\\
\hline
MiSCSD	&0.595	&0.7596\\
\hline
VWHD	&0.5945	&0.7428\\
\hline
JacD	&0.5936	&0.746\\
\hline
DicD	&0.5936	&0.746\\
\hline
ED 	&0.5927	&0.7429\\
\hline
SED	&0.5927	&0.7429\\
\hline
AD	&0.5927	&0.7429\\
\hline
KD	&0.5911	&0.7154\\
\hline
MCED	&0.5889	&0.7416\\
\hline
HamD	&0.5843	&0.7048\\
\hline
CD	    &0.5742	&0.7154\\
\hline
HauD	&0.5474	&0.621\\
\hline
KDD	    &0.5295	&0.5357\\
\hline
TopD	&0.5277	&0.6768\\
\hline
JSD	    &0.5276	&0.6768\\
\hline
CSSD	&0.5273	&0.4895\\
\hline
KLD	    &0.5089	&0.5399\\
\hline
MeeD	&0.4958	&0.4324\\

\hline
BD	    &0.4747	&0.4908\\
\hline
   \end{tabular}
  }
 \end{center}
\end{table}

\section{Conclusions and Future Perspectives}\label{sec:conc}

In this review, the performance (accuracy, precision and recall) of the KNN classifier has been evaluated using a large number of distance measures, on clean and noisy datasets, attempting to find the most appropriate distance measure that can be used with the KNN in general. In addition we tried finding the most appropriate and robust distance that can be used in the case of noisy data.
A large number of experiments conducted for the purposes of this review, and the results and analysis of these experiments show the following:   
\begin{enumerate}
	\item	The performance of KNN classifier depends significantly on the distance used, the results showed large gaps between the performances of different distances. For example we found that Hassanat distance performed the best when applied on most datasets comparing to the other tested distances.
	\item	We get similar classification results when we use distances from the same family having almost the same equation, some distances are very similar, for example, one is twice the other, or one is the square of another. In these cases, and since the KNN compares examples using the same distance, the nearest neighbors will be the same if all distances were multiplied or divided by the same constant.
	\item	There was no optimal distance metric that can be used for all types of datasets, as the results show that each dataset favors a specific distance metric, and this result complies with the no-free-lunch theorem.
	\item	The performance (measured by accuracy, precision and recall) of the KNN degraded only about $20\%$ while the noise level reaches $90\%$, this is true for all the distances used. This means that the KNN classifier using any of the top $10$ distances tolerate noise to a certain degree. 
\item	Some distances are less affected by the added noise comparing to other distances, for example we found that Hassanat distance performed the best when applied on most datasets under different levels of heavy noise.
 \end{enumerate}

Our work has the following limitations, and future works will focus on studying, evaluating and investigating these.  
\begin{enumerate}
	\item	Although, we have tested a large number of distance measures, there are still other distances and similarity measures that are available in the machine learning area that need to be tested and evaluated for optimal performance with and without noise.
	\item	The $28$ datasets though higher than previously tested, still might not be enough to draw significant conclusions in terms of the effectiveness of certain distance measures, and therefore, there is a need to use a larger number of datasets with varied data types. 
	\item The creation of noisy data is done by replacing a certain percentage (in the range $10\%$ - $90\%$) of the examples by completely random values in the attributes. We used this type of noise for its simplicity and straightforwardness in the interpretation of the effects of distance measure choice with KNN classifier. However, this type of noise might not simulate other types of noise that occur in the real-world data. It is an interesting task to try other realistic noise types to evaluate the distance measures for robustness in a similar manner.
	\item	Only KNN classifier was reviewed in this work, other variants of KNN such as~\citep{Hassanat2018b, Hassanat2018c, Hassanat2018d, Hassanat2018a} need to be investigated.
	\item	Distance measures are not used only with the KNN, but also with other machine learning algorithms, such as different types of clustering, those need to be evaluated under different distance measures.
\end{enumerate}


\end{document}